\documentclass{article}

\usepackage{microtype}
\usepackage{graphicx}
\usepackage{subfigure}
\usepackage{booktabs} 

\usepackage{hyperref}

\usepackage[square,numbers]{natbib} 
\usepackage{adjustbox} 
\usepackage{tikz} 
\usetikzlibrary{arrows,shapes,positioning,automata}
\usetikzlibrary{calc,decorations.markings}
\usetikzlibrary{decorations.pathreplacing}
\usepackage{amsmath} 
\usepackage{empheq}
\usepackage{algorithmic}  
\usepackage[algo2e,ruled,vlined,noend]{algorithm2e} 
\usepackage{multirow,multicol}
\usepackage{amsfonts}
\usepackage{amsmath}

\usepackage{stfloats} 
\usepackage[justification=centering]{caption} 
\usepackage{empheq} 
\usepackage[most]{tcolorbox}

\newcommand{\RNum}[1]{\uppercase\expandafter{\romannumeral #1\relax}}

\usepackage[accepted]{icml2021}

\icmltitlerunning{A Practical Layer-Parallel Training Algorithm for Residual Networks}

\begin{document}

\twocolumn[
\icmltitle{A Practical Layer-Parallel Training Algorithm for Residual Networks}



\icmlsetsymbol{equal}{*}

\begin{icmlauthorlist}
\icmlauthor{Qi Sun}{equal,to}
\icmlauthor{Hexin Dong}{equal,to}
\icmlauthor{Zewei Chen}{equal,goo}
\icmlauthor{Weizhen Dian}{to}
\icmlauthor{Jiacheng Sun}{goo}
\icmlauthor{Yitong Sun}{goo}
\icmlauthor{Zhenguo Li}{goo}
\icmlauthor{Bin Dong}{to}
\end{icmlauthorlist}

\icmlaffiliation{to}{Peking University}
\icmlaffiliation{goo}{Huawei Noah's Ark Lab}

\icmlcorrespondingauthor{Qi Sun}{qsun2019@pku.edu.cn}
\icmlcorrespondingauthor{Bin Dong}{dongbin@math.pku.edu.cn}

\icmlkeywords{Machine Learning, ICML}

\vskip 0.3in
]



\printAffiliationsAndNotice{\icmlEqualContribution} 

\begin{abstract}
Gradient-based algorithms for training ResNets typically require a forward pass of the input data, followed by back-propagating the objective gradient to update parameters, which are time-consuming for deep ResNets. To break the dependencies between modules in both the forward and backward modes, auxiliary-variable methods such as the penalty and augmented Lagrangian (AL) approaches have attracted much interest lately due to their ability to exploit layer-wise parallelism. However, we observe that large communication overhead and lacking data augmentation are two key challenges of these methods, which may lead to low speedup ratio and accuracy drop across multiple compute devices. Inspired by the optimal control formulation of ResNets, we propose a novel serial-parallel hybrid training strategy to enable the use of data augmentation, together with downsampling filters to reduce the communication cost. The proposed strategy first trains the network parameters by solving a succession of independent sub-problems in parallel and then corrects the network parameters through a full serial forward-backward propagation of data. Such a strategy can be applied to most of the existing layer-parallel training methods using auxiliary variables. As an example, we validate the proposed strategy using penalty and AL methods on ResNet and WideResNet across MNIST, CIFAR-10 and CIFAR-100 datasets, achieving significant speedup over the traditional layer-serial training methods while maintaining comparable accuracy.
\end{abstract}

\section{Introduction}

Deep neural networks with millions of trainable parameters have become indispensable tools for machine learning applications involving large datasets \cite{lecun2015deep}. For the solution of such large-scale optimization problems, gradient-based algorithms are often employed, which requires a forward pass of the input data, followed by the backpropagation \cite{rumelhart1985learning} of objective gradients to update parameters in each iteration step. However, even with the use of modern Graphical Processing Units (GPUs), the overall training process still remains time-consuming. As such, various parallelization techniques including, but not limited to data-parallelism \cite{iandola2016firecaffe}, model-parallelism \cite{dean2012large}, and a combination of both \cite{paine2013gpu,harlap2018pipedream} have been proposed to reduce the training runtimes. Unfortunately, none of the above methods could tackle the scalability barrier created by the intrinsically serial propagation of data within the network \cite{gunther2020layer}, preventing us from updating layers in parallel and fully leveraging the computing resources.

One way of achieving speed-up over the traditional methods is to apply synthetic gradients to build decoupled neural interfaces \cite{jaderberg2017decoupled}, where the objective gradients are approximated by additional neural networks so that each layer can be locally updated without performing the full serial backpropagation. However, it fails in training deep convolutional networks since the construction of synthetic loss function has little relation to the target objective function \cite{miyato2017synthetic}. Another related work is proposed in \cite{huo2018decoupled}, where the authors use stale gradients to remove the backward locking. Such a method, though effective, requires a full serial forward pass before executing the decoupled parallel backpropagation, thereby the upper bound of the resulting speed-up is very limited according to Amdahl's law \cite{gustafson1988reevaluating}. 

To exploit layer-wise parallelism in both the forward and backward modes, several algorithms were proposed recently by introducing auxiliary variables associated with the decoupled neural interfaces, \textit{e.g.}, the quadratic penalty method \cite{carreira2014distributed,zeng2018global,choromanska2018beyond}, the augmented Lagrangian (AL) method \cite{taylor2016training,zeng2019convergence,marra2020local} and the proximal method \cite{li2020training} for training fully-connected networks, which can achieve speed-up over the traditional layer-serial training method on a single Central Processing Unit (CPU) \cite{li2020training}. However, as shown in \cite{gotmare2018decoupling} and commented in \cite{huo2018decoupled}, the performances of most of these methods are much worse than the backpropagation algorithm for deep convolutional neural networks. In other recent approachs \cite{gunther2020layer,parpas2019predict,kirby2020layer}, based on the similarity of ResNets training to the optimal control of nonlinear systems \cite{weinan2017proposal}, parareal method for solving differential equations is employed to replace the forward pass and backpropagation with iterative multigrid schemes respectively. Since the feature maps need to be recorded at each module and then used in a subsequent process to solve the adjoint equation \cite{troltzsch2010optimal}, experiments were conducted on simple ResNets across small datasets, rather than state-of-the-art ResNets across larger datasets. So far, to the best of our knowledge, it is uncertain that whether these layer-parallel training strategies can be effectively and efficiently applied to modern deep networks across real-world datasets.

In this work, we observe that there are two key issues that prevent us from attaining good performance in practical scenarios. The accuracy drop of trained model is mainly due to the lack of data augmentation, which is hard to implement at the presence of auxiliary variables. Furthermore, data communication is another potential issue that may hamper the speed-up ratio, which was not adequately addressed in previous studies since most implementations were conducted on CPUs. Based on these observations and inspired by optimal control formulation of training ResNets, we propose a novel serial-parallel hybrid (SPH) training strategy that alternates between the traditional layer-serial training with data augmentation and the layer-parallel training in a reduced parameter space. Here, layer-serial training allows the use of data augmentation while layer-parallel training in a reduced parameter space is achieved by downsampling (DS) of the auxiliary variables to alleviate the communication burden.

The contribution of this work is threefold:

\vspace{-0.25cm}

\begin{itemize}
\setlength\itemsep{0.05em}
\item[$(1)$] We observe that large communication overhead and the lack of data augmentation are two key challenges for auxiliary-variable methods, which may lead to accuracy drop and low speedup ratio across multiple computing devices.
\item[$(2)$] A novel SPH strategy is proposed to enable the use of data augmentation during training, together with the employment of downsampling filters to reduce the communication cost.
\item[$(3)$] We validate our methods on ResNet and WideResNet across MNIST, CIFAR-10, and CIFAR-100 datasets, achieving significant speed-up over the traditional layer-serial training methods while maintaining comparable accuracy.
\end{itemize}

\vspace{-0.25cm}

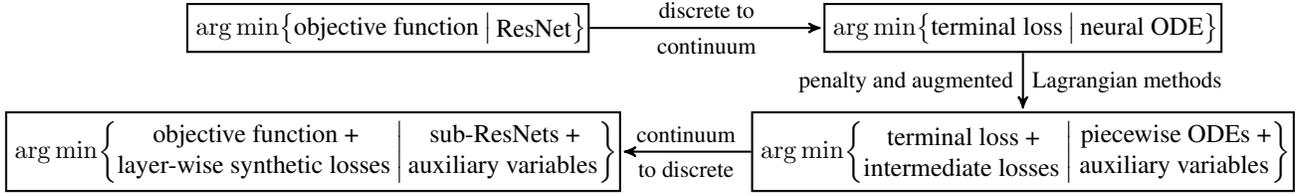
\begin{figure*}[t]
\begin{adjustbox}{max totalsize={\textwidth}{\textheight},center}
\begin{tikzpicture}[shorten >=1pt,auto,node distance=3.5cm,thick,main node/.style={rectangle,draw,font=\normalsize},decoration={brace,mirror,amplitude=7},every text node part/.style={align=center}]
\node[main node] (1) {\parbox{5.26cm}{\centering $\operatorname*{arg\,min}\!\left\{ \textnormal{objective function} \,\big|\, \parbox{1.03cm}{\centering ResNet} \right\}$}};
\node[main node] (2) [right =3.2cm of 1] {\parbox{5.2cm}{\centering $\operatorname*{arg\,min}\!\left\{ \textnormal{terminal loss} \,\big|\, \textnormal{neural ODE} \right\}$}};   
\node[main node] (3) [below =0.8cm of 2] {\parbox{7.18cm}{\centering $\operatorname*{arg\,min}\!\left\{ \parbox{2.7cm}{\centering terminal loss + \\ intermediate losses} \,\bigg|\, \parbox{2.65cm}{\centering piecewise ODEs + \\ auxiliary variables} \right\}$}};    
\node[main node] (4) [left =1.78cm of 3] {\parbox{8.17cm}{\centering $\operatorname*{arg\,min}\!\left\{ \parbox{3.75cm}{\centering objective function + \\ layer-wise synthetic losses} \,\bigg|\, \parbox{2.6cm}{\centering sub-ResNets + \\ auxiliary variables } \right\}$}};
\path[every node/.style={font=\sffamily\small},->,>=stealth']
	(1) edge node[above,midway] {\textnormal{discrete to}} node[below,midway] {\textnormal{continuum}} (2)
	(2) edge node[left,midway] {\textnormal{penalty and augmented}} node[right,midway] {\textnormal{Lagrangian methods}} (3)    
    (3) edge node[above,midway] {\textnormal{continuum}} node[below,midway] {\textnormal{to discrete}} (4);   
\end{tikzpicture}
\end{adjustbox}
\vspace{-0.6cm}
\caption{A diagram describing the construction of parallel training algorithm from the dynamical systems view.}
\label{fig-big-picture}
\end{figure*}

The rest of this paper is organized as follows. Section 2 is devoted to recalling the backpropagation algorithm and locking effects during training, followed by the layer-parallel training of ResNets from a dynamical systems viewpoint (see \autoref{fig-big-picture}). The downsampling filters and SPH strategy are proposed in Section 3 to handle the issues of data communication and data augmentation. Experimental results are presented in Section 4 to validate our theoretical findings.

\begin{figure*}[b]
\begin{adjustbox}{max totalsize={\textwidth}{\textheight},center}
\begin{tikzpicture}[xscale=1]
\draw [black] [thick] (0.8,-0.2) -- (10.4,-0.2);
\foreach \x in {0,...,24}
      \draw [black] [thick] (0.8+0.4*\x,-.3) -- (0.8+0.4*\x,-.1);
\node[above] at (0.5,-.1) {$0=t_0$};
\node[above] at (1.6,-.1) {$\cdots$};
\node[above] at (2.4,-.1) {$t_n$};
\node[above] at (3.2,-.1) {$\cdots$};
\node[above] at (4,-.1) {$\cdots$};
\node[above] at (4.8,-.1) {$\cdots$};
\node[above] at (5.6,-.1) {$t_{kn}$};
\node[above] at (6.4,-.1) {$\cdots$};
\node[above] at (7.2,-.1) {$t_{kn+n}$};
\node[above] at (8,-.1) {$\cdots$};
\node[above] at (8.8,-.1) {$\cdots$};
\node[above] at (9.6,-.1) {$\cdots$};
\node[above] at (10.7,-.1) {$t_L=1$};

\draw [-,thick, blue] (0.8,-2) to [out=60,in=160] (4,-1.2)
to [out=340,in=160] (7.2,-1) to [out=340,in=220] (10.4,-0.6);

\draw [black] [thick] (0.8,-2.5) -- (10.4,-2.5);
\foreach \x in {0,...,6}
      \draw [fill=black] [black] [thick] (0.8+1.6*\x,-2.5) circle [radius=.05];
\node[below] at (0.5,.-2.55) {$0=s_0$};
\node[below] at (2.4,.-2.55) {$s_1$};
\node[below] at (4,.-2.55) {$\cdots$};
\node[below] at (5.6,.-2.55) {$s_k$};
\node[below] at (7.2,.-2.55) {$s_{k+1}$};
\node[below] at (8.8,-2.55) {$\cdots$};
\node[below] at (10.7,.-2.55) {$s_K=1$};

\draw [fill=blue] [blue] [thick] (0.8,-2) circle [radius=.05];
\draw [fill=red] [blue] [thick] (2.4,-0.98) circle [radius=.05];
\draw [red] [thick] (2.4,.-0.5) circle [radius=.05];
\draw [-,thick,red] (0.8,-1.95) to [out=90,in=200] (2.35,.-0.5);

\draw [fill=blue] [blue] [thick] (4,.-1.2) circle [radius=.05];
\draw [fill=red] [red] [thick] (2.4,-1.6) circle [radius=.05];
\draw [red] [thick] (4,.-2.1) circle [radius=.05];
\draw [-,thick,red] (2.4,-1.6) to [out=60,in=200] (4,.-2.05);
\draw [black] [dotted,thick] (2.4,-0.56) -- (2.4,-0.92) ;
\draw [black] [dotted,thick] (2.4,-1.03) -- (2.4,-1.55) ;
\draw [black] [dotted,thick] (4,-0.55) -- (4,-1.15) ;
\draw [black] [dotted,thick] (4,-2.05) -- (4,-1.23) ;

\draw [fill=red] [red] [thick] (4,-0.6) circle [radius=.05];
\draw [red] [thick] (5.6,.-1.95) circle [radius=.05];
\draw [-,thick,red] (4,-0.6) to [out=340,in=200] (5.6,.-1.9);

\draw [fill=blue] [blue] [thick] (5.6,.-1.1) circle [radius=.05];
\draw [fill=blue] [blue] [thick] (7.2,.-1) circle [radius=.05];
\draw [fill=red] [red] [thick] (5.6,.-1.65) circle [radius=.05];
\draw [-,thick,red] (5.6,.-1.65) to [out=340,in=180] (7.2,.-0.55);
\draw [red] [thick] (7.2,.-0.6) circle [radius=.05];
\draw [black] [dotted,thick] (5.6,.-1.6) -- (5.6,.-1.15) ;
\draw [black] [dotted,thick] (5.6,.-1.9) -- (5.6,.-1.7) ;

\draw [fill=blue] [blue] [thick] (8.8,.-1.27) circle [radius=.05];
\draw [fill=red] [red] [thick] (7.2,.-1.65) circle [radius=.05];
\draw [red] [thick] (8.8,.-1) circle [radius=.05];
\draw [-,thick,red] (7.2,.-1.65) to [out=340,in=180] (8.75,.-1);
\draw [black] [dotted,thick] (7.2,.-1.6) -- (7.2,.-1.05);
\draw [black] [dotted,thick] (7.2,.-0.65) -- (7.2,.-0.95) ;

\draw [fill=blue] [blue] [thick] (10.4,.-0.6) circle [radius=.05];
\draw [fill=red] [red] [thick] (8.8,.-1.8) circle [radius=.05];
\draw [red] [thick] (10.4,.-1) circle [radius=.05];
\draw [-,thick,red] (8.8,.-1.8) to [out=40,in=260] (10.4,.-1.05);
\draw [black] [dotted,thick] (8.8,.-1.32) -- (8.8,.-1.8);
\draw [black] [dotted,thick] (8.8,.-1.06) -- (8.8,.-1.22);
\draw [black] [dotted,thick] (10.4,.-0.65) -- (10.4,.-0.95);
\end{tikzpicture}
\end{adjustbox}
\vspace{-0.6cm}
\caption{Contrary to the trajectory of neural ODE (blue line), introducing auxiliary variables (solid red dots) for each local sub-problem (red lines) enables a time-parallel computation of the state, adjoint, and control variables. Note that to approximately recover the serial approach, violation of equality constraints (mismatch between solid and hollow red dots) should be penalized in the objective function.}
\label{fig-multifidelity-forward-propagation}
\end{figure*}
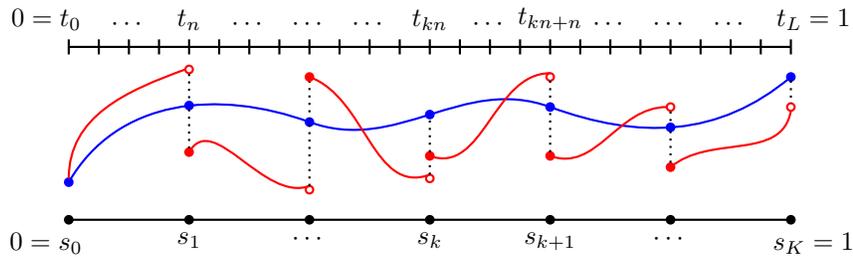

\section{Preliminaries}

\subsection{Layer-Serial Training} \label{section-ResNet}

Based on the concept of modified equations \cite{weinan2017proposal} or the variational analysis using $\Gamma$-convergence \cite{thorpe2018deep}, training of the ResNets from scratch \cite{he2016deep,he2016identity} (see \autoref{appendix-ResNet} for notation description)
\begin{equation}
	\operatorname*{arg\,min}_{\{W_\ell\}_{\ell=0}^{L-1}} \left\{ \varphi(X_L)  \, \Big| \, X_0\!=\!S(y), \, X_{\ell+1} \!=\! X_\ell \!+\! F(X_\ell,W_\ell) \right\} 
	\label{ResNet-Training-Task}
\end{equation}
can be interpreted as the discretization of a terminal control problem governed by the so-called neural ODE \cite{chen2018neural}
\begin{equation}
	\operatorname*{arg\,min}_{\omega_t} \left\{ \varphi(x_1) \, \Big| \, x_0=S(y) , \, dx_t = f(x_t,w_t)dt \right\}.
	\label{ODE-Training-Task}
\end{equation}
For the ease of comparison, we refer, respectively, to \eqref{ResNet-Training-Task} and \eqref{ODE-Training-Task} as the layer-serial and time-serial training method, and the same is said for their variants in the following sections.

Moreover, given a learning rate $\eta>0$, the continuous-time counterpart of the backpropagation algorithm \cite{hecht1992theory} for solving \eqref{ResNet-Training-Task}, \textit{i.e.}, for $0\leq \ell\leq L-1$,
\begin{equation}
	W_\ell \leftarrow W_\ell - \eta \frac{\partial \varphi(X_L)}{\partial X_{\ell+1}}\frac{\partial F(X_{\ell},W_\ell)}{\partial W_\ell}, 
	\label{ResNet-BackPropagation}
\end{equation}
is handled by the adjoint and control equations for finding the extremal of \eqref{ODE-Training-Task} \cite{li2017maximum}, that is,
\begin{empheq}[left = ]{align}
	\displaystyle & \ dp_t = -p_t \frac{\partial f(x_t,w_t)}{\partial x}dt, & & p_1=\frac{\partial \varphi(x_1)}{\partial x},  \label{ODE-Adjoint-Equation}\\
    \displaystyle & \ w_t \leftarrow w_t - \eta p_t \frac{\partial f(x_t,w_t)}{\partial w}, & &  0\leq t\leq 1, \label{ODE-Weights-Update}
\end{empheq}
where $p_t$ is the adjoint variable that captures the objective changes with respect to hidden neurons (see \autoref{appendix-ResNet}).

As a result, the $locking$ effects \cite{jaderberg2017decoupled} for training feedforward neural networks, \textit{i.e.},

\vspace{-0.25cm}

\begin{itemize}
\setlength\itemsep{0.05em}
\item[(i)] forward locking: no module can process its incoming data before the previous node in the directed forward network have executed;
\item[(ii)] backward locking: no module can capture the objective changes with respect to its activation layer before the previous node in the backward network have executed;
\item[(iii)] update locking: no module parameters can be updated before all the dependent nodes have executed in both the forward and backward modes;
\end{itemize}

\vspace{-0.25cm}

can be recast as the necessity of solving both the neural ODE in \eqref{ODE-Training-Task} and a backward-in-time adjoint equation \eqref{ODE-Adjoint-Equation} in order to perform the control updates \eqref{ODE-Weights-Update}.

This connection not only brings us a dynamical system view of the locking effects but also provides a way to consistently discretize the iterative system \eqref{ODE-Adjoint-Equation} and \eqref{ODE-Weights-Update} for solving the continuous-time optimization problem \eqref{ODE-Training-Task}. Although recent hardware developments have gradually increased the capability of data-parallelism \cite{iandola2016firecaffe}, model-parallelism \cite{dean2012large}, and a combination of both \cite{paine2013gpu,harlap2018pipedream} for training large-scale neural networks, none of them could overcome the scalability barrier caused by the serial forward-backward propagation of data through the network \cite{gunther2020layer}. As such, breaking the locking issues, or, equivalently, parallelizing the iterative system for solving \eqref{ODE-Training-Task} is another promising approach to speed up the training.

\subsection{Layer-Parallel Training} \label{section-parareal-terminal-control}

The similarity of training ResNets to the terminal control of ODEs motivates us to use the parallel-in-time methods \cite{maday2002parareal,carraro2015multiple} to achieve concurrency across all the network modules (see \autoref{fig-big-picture}). 

\subsubsection{Forward Pass with Auxiliary Variables}

That is, to employ $K\in\mathbb{N}_+$ independent processors for the solution of neural ODE in \eqref{ODE-Training-Task}, we introduce a partition of $[0,1]$ into several disjoint intervals as shown in \autoref{fig-multifidelity-forward-propagation}
\begin{equation*}
    0 = s_0  < \ldots < s_k < s_{k+1} < \ldots < s_{K}=1.
\end{equation*}

Now we are ready to define the piecewise states $\{x_t^k\}_{k=0}^{K-1}$ such that the underlying dynamic evolves according to 
\begin{equation}
	x^k_{s_k^+}=\lambda_k, \qquad  d x^k_t = f(x^k_t,w^k_t)dt\ \ \ \textnormal{on}\ (s_k,s_{k+1}],
	\label{Parallel-ODE-State-Equation}
\end{equation}
\textit{i.e.}, the continuous-time forward pass that originates from auxiliary variable $\lambda_k$ and with control variable $w_t^k$. Here, $x^k_{s_k^+}$ and $x^k_{s_k^-}$ refer to the right and left limits of the possibly discontinuous function $x^k_t$ at the interface $t=s_k$. 

Clearly, the state variable of problem \eqref{ODE-Training-Task} satisfies $x_t = x^k_t$ for any $t\in[s_k,s_{k+1}]$ and $0\leq k\leq K-1$ if and only if
\begin{equation*}
	w_t^k = w_t |_{(s_k,s_{k+1})} \qquad \textnormal{and} \qquad \lambda_k = x_{s_k^+},
\end{equation*}
or, equivalently, $\lambda_k = x^{k-1}_{s_{k}^-}$ with $x^{-1}_{s_{0}^-}=x_0$ to replace the second condition. As a result, the optimization problem \eqref{ODE-Training-Task} can be reformulated as
\begin{equation}
\begin{gathered}
	\operatorname*{arg\,min}_{\{w^k_t\}_{k=0}^{K-1}} \left\{ \varphi(x^{K-1}_{s_K^-})  \, \Big| \, x^{k-1}_{s_{k}^-}=\lambda_k \ \ \text{and}\ \ \eqref{Parallel-ODE-State-Equation} \right\}
\end{gathered}
	\label{ODE-Training-Task-ReWritten}
\end{equation}
which offers the possibility of parallelizing the evolution of dynamical system \eqref{Parallel-ODE-State-Equation} by relaxing the other constraint, \textit{e.g.}, \autoref{fig-multifidelity-forward-propagation} with external auxiliary variables. 

\begin{figure*}[t]
\vspace{-0.2cm}
\centering
\includegraphics[width=0.98\textwidth]{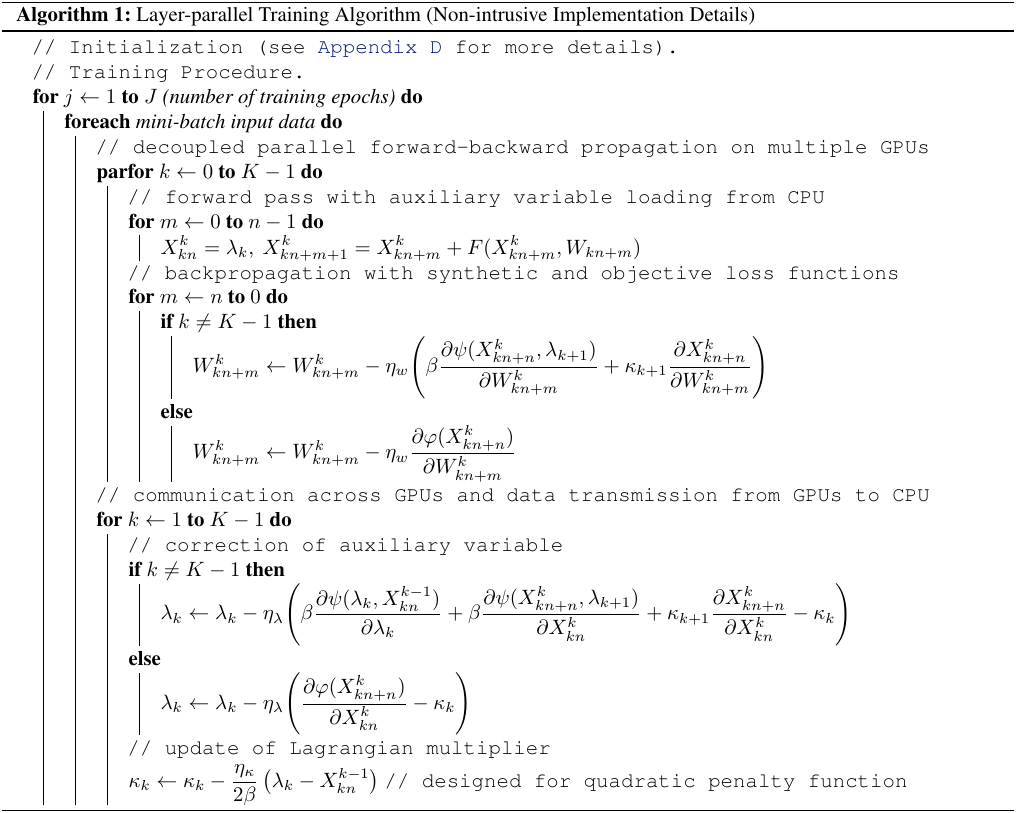}
\vspace{-0.1cm}
\end{figure*}

\subsubsection{Augmented Lagrangian Method}

Formula \eqref{ODE-Training-Task-ReWritten} implies that the exact connection between adjacent intervals can be loosened by incorporating external auxiliary variables, which inspires us to  relax the equality constraints and add penalties to the objective function, \textit{i.e.},
\begin{equation}
	\operatorname*{arg\,min}_{\{w_t^k,\lambda_k\}_{k=0}^{K-1}} \bigg\{ \varphi(x^{K-1}_{s_K^-}) + \beta \sum_{k=0}^{K-1} \psi(\lambda_k,x^{k-1}_{s_{k}^-}) \, \Big| \, \eqref{Parallel-ODE-State-Equation}\bigg\}
	\label{Parallel-ODE-Training-Task}
\end{equation}
where $\beta>0$ is a scalar constant and $\psi(\lambda,x)= \| \lambda - x\|_{\ell_2}^2$ the quadratic penalty function. Such a method has been extensively used due to its simplicity and intuitive appeal \cite{taylor2016training,choromanska2018beyond,gotmare2018decoupling}, however, it suffers from ill-conditioning when the penalty coefficient is large \cite{nocedal2006numerical}. 

To make the approximate solution of \eqref{Parallel-ODE-Training-Task} nearly satisfy the layer-serial approach \eqref{ODE-Training-Task} even for moderate values of $\beta$, we consider the augmented Lagrangian of problem \eqref{ODE-Training-Task-ReWritten} 
\begin{empheq}[left = ]{align*}
	\displaystyle & \mathcal{L}( x_t^k, p_t^k, w^k_t, \lambda_k, \kappa_k ) =  \varphi(x^{K-1}_{s_K^-})  +  \sum_{k=0}^{K-1} \Bigg( \beta \psi(\lambda_k,x^{k-1}_{s_{k}^-}) \\
    \displaystyle & \qquad \ + \int_{s_k}^{s_{k+1}} p^k_t \big( f(x_t^k,w^k_t) - \dot{x}_t^k \big)dt - \kappa_k( \lambda_k - x^{k-1}_{s_{k}^-} ) \Bigg)
\end{empheq}
where $\kappa_k$ denotes an explicit Lagrange multiplier. Notably, by forcing $\kappa_k\equiv 0$ for any $0\leq k\leq K-1$, the augmented Lagrangian method degenerates the penalty approach.

To clarify the differences between layer-parallel training of fully-connected networks \cite{zeng2019convergence} and ResNets, we refer the readers to \autoref{fig-big-picture} for  technical details. Specifically, the iterative system for solving the relaxed optimization problems is provided in \autoref{appendix-Calculsu-Variations}, which results in a non-intrusive layer-parallel training algorithm after employing the consistent discretization schemes discussed in \autoref{appendix-ResNet}. Implementation details are summarized in Algorithm 1, which includes both the AL and penalty (achieved by forcing $\kappa_k\equiv 0$ during training) methods.


\begin{figure}[t]
\begin{adjustbox}{center}
\begin{tikzpicture}[shorten >=1pt,auto,node distance=3.5cm,thick,main node/.style={rectangle,draw,font=\normalsize},decoration={brace,mirror,amplitude=7},every text node part/.style={align=center}]
\node[main node] (1) {\parbox{3.8cm}{\centering \parbox{3.8cm}{\centering layer-parallel training \\ in reduced parameter space }}};
\node[] (2) [below =0.01cm of 1] {$\times\ m$ times};
\node[main node] (3) [right =0.5cm of 1] {\parbox{3.3cm}{\centering $\parbox{3.3cm}{\centering layer-serial training \\ with data augmentation}$}};
\node[] (4) [below =0.01cm of 3] {$\times\ n$ times};

\path[every node/.style={font=\sffamily\small},->,>=stealth']	
	(1) edge  (3);
	   
\end{tikzpicture}
\end{adjustbox}
\vspace{-0.8cm}
\caption{A diagram describes the serial-parallel hybrid strategy.}
\label{fig-SPH}
\end{figure}
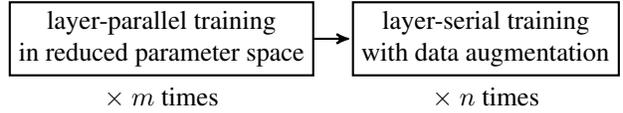

\section{Method}

We note that the external auxiliary variables increases concurrency across all the network modules but incurs additional memory and communication overheads, which may limit the performance for the exposed parallelism especially in the setting of fine partitioned models with data augmentation. Therefore, a novel method with serial-parallel hybrid training strategy and downsampling strategy is proposed to handle these issues (see Figure 3 for the schematic of the proposed strategy). We also note that our strategy is effective on any layer-parallel training algorithm that introduces auxiliary variables.

\subsection{Downsampling for Data Communication}

For each iteration of Algorithm 1, the computational time associated with the layer-serial ($K=1$) and layer-parallel ($K>1$) training approaches can be summarized as follows:
\begin{center}
\begin{small}
\begin{tabular}{ccc}
\toprule
 &  layer-serial & layer-parallel  \\
\midrule
forward pass & $t_f$ &  $\frac{1}{K}t_f$  \\
backpropagation & $t_b$ & $\frac{1}{K}t_b + t_\psi $ \\
communication & $t_d$ & $t_\lambda$ + $t_\kappa$ \\
\bottomrule
\end{tabular}
\end{small}
\end{center}
where $t_f$ $(t_b)$ denotes the time cost of forward pass (backpropagation) using the layer-serial training method, $t_d$ the time cost on data loader, $t_\psi$ the computation time of synthetic loss functions, $t_\lambda + t_\kappa$ the evaluation and communication time of auxiliary variables. Clearly, the speedup ratio per epoch can be expressed as
\begin{equation}
	\rho_K = \frac{\text{serial runtime}}{\text{parallel runtime}}  = \frac{1}{ \frac{1}{K} \frac{t_f+t_b}{t_f+t_b+t_d}  + \frac{t_\psi + t_\lambda + t_\kappa}{t_f+t_b+t_d} }
	\label{speedup-ratio}
\end{equation}
where $t_f$, $t_b$, $t_d$, $t_\psi$, $t_\lambda$ and $t_\kappa$ are almost independent of the model partition number $K$ during training.

Note that for realistic neural networks such as ResNets \cite{he2016deep,he2016identity}, it is plausible to assume that $t_f+t_b+t_d> t_\psi + t_\lambda + t_\kappa$, which immediately shows speed-up over the traditional layer-serial training approach by choosing a sufficient large value of $K$. Notably, formula \eqref{speedup-ratio} also implies that the upper bound of speed-up ratio is given by
\begin{equation*}
	\rho_K < \frac{t_f+t_b+t_d}{t_\psi + t_\lambda + t_\kappa},
\end{equation*}
namely, the communication becomes the performance bottleneck as the model is partitioned more
finely, which motivates us to reduce the data communication overhead in order to further accelerate the network training.


One way to achieve this is to design downsampling (DS) filters to attenuate the size of auxiliary variables. We can, for instance, take the example of penalty method \eqref{Parallel-ODE-Training-Task}. Instead of transferring the full-size auxiliary variables between CPU and GPU cores, we can operate with the downsampled data
\begin{equation*}
	\Lambda_k = \textnormal{DS}(\lambda_k),\ \  \textnormal{or approximately,} \ \ \lambda_k \approx \textnormal{US}(\Lambda_k)
\end{equation*}
to execute the forward pass \eqref{Parallel-ODE-State-Equation} for $0\leq k\leq K-1$
\begin{equation}
	x^k_{s_k^+}=\textnormal{US}(\Lambda_k), \ \ d x^k_t = f(x^k_t,w^k_t)dt\ \ \textnormal{on}\ (s_k,s_{k+1}].
	\label{Parallel-ODE-State-Equation-DS}
\end{equation}
For instance, by taking the Kronecker product with an all-ones matrix of size $2\times 2$ for each slice of the tensor $\Lambda_k$, we obtain the auxiliary variable $\lambda_k$ for forward pass. \autoref{section-experiments} will focus on this particular example and we leave the exploration of other downsampling tools as future work.

As such, the optimization problem is now defined in a reduced parameter space, that is,
\begin{equation*}
	\operatorname*{arg\,min}_{\{w_t^k,\Lambda_k\}_{k=0}^{K-1}} \bigg\{ \varphi(x^{K-1}_{s_K^-}) + \beta \sum_{k=0}^{K-1} \psi(\textnormal{US}(\Lambda_k),x^{k-1}_{s_{k}^-}) \, \Big| \, \eqref{Parallel-ODE-State-Equation-DS}\bigg\}.
	\label{Parallel-ODE-Training-Task-DS}
\end{equation*}

With a slight loss of accuracy, the memory and communication overheads can be significantly reduced compared with the original method \eqref{Parallel-ODE-Training-Task}. Moreover, the implementation is very straightforward, only requiring an additional upsampling layer before the execution of forward pass in Algorithm 1, while the backpropagation is automatically achieved through the standard auto-differentiation. Such a technique can also be easily extended to the augmented Lagrangian method (omitted here for simplicity).

\subsection{Hybrid Training for Data Augmentation}


Another key observation is that each training sample requires to introduce a group of corresponding auxiliary variables. These auxiliary variables need to be stored and recomputed during the iteration. When cooperating with the commonly used data augmentation \cite{tanner1987calculation} which reduce overfitting by artificially increasing the number of samples in the training set, extra auxiliary variables need to introduce for each augmented sample. It would incur prohibitive memory requirements making great challenge to incorporate data augmentation. We believe that this is the key reason that previous methods' performance is far below the state-of-the-art.

\begin{figure}[t]
\centering
\includegraphics[width=0.75\columnwidth]{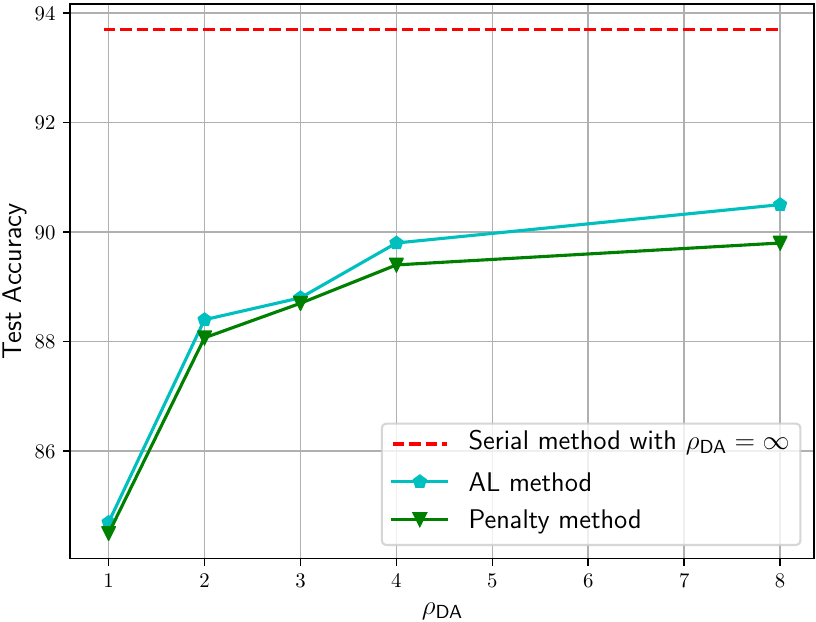}
\vspace{-0.2cm}
\caption{Test accuracy of trained model using data augmentation.}
\label{figure-DAR}
\end{figure}

To justify our argument, we shown some preliminary experimental results by using different ratios $\rho_{\textnormal{DA}}$ of data augmentation (\textit{i.e.}, the number of synthetic images to the number of real images), \autoref{figure-DAR} shows the test accuracy of ResNet-110 for the classification task on CIFAR-10 dataset, where $\rho_{\textnormal{DA}}=\infty$ denotes the data augmentation containing random operations. It can be observed that, as $\rho_{\textnormal{DA}}$ is increased, the accuracy gap between the traditional layer-serial training method and the proposed layer-parallel training approach is tending to close. However, the memory requirements for storing all the synthetic training data blows up 
even for moderate values of $\rho_{\textnormal{DA}}$, which is unaffordable in practical scenarios.

Unfortunately, most of the existing auxiliary-variable methods fail to address this issue, which often leads to a significant accuracy drop of the trained networks \cite{gotmare2018decoupling}. To allow the use of data augmentation during training, we propose a novel serial-parallel hybrid (SPH) strategy that alternatives between the layer-serial and layer-parallel training modes as depicted in \autoref{fig-SPH}. That is, the layer-parallel training method is performed $m$ times without data augmentation, followed by $n$ times execution of the layer-serial training method, which improves the network parameters through the employment of data augmentation. 

As an immediate result, the speed-up ratio now gives
\begin{equation}
	\rho_H = \frac{(m+n)\times t_s}{m \times t_p + n \times t_s} = \frac{1+\gamma_H}{\frac{1}{\rho_K}+\gamma_H} 
	\label{speedup-ratio-SPH}
\end{equation}
where $\gamma_H=\frac{n}{m}$ indicates the hybrid ration, $t_s=t_f+t_b+t_d$ and $t_p=\frac{1}{K}(t_s+t_b)+t_\psi+t_\lambda+t_\kappa$ are runtime of the layer-serial and layer-parallel training methods per epoch. Although the serial portions in \eqref{speedup-ratio-SPH} hamper the speedup ratio, \textit{i.e.}, $\rho_H < \rho_K$, enabling data augmentation can significantly increase the test accuracy as shown in \autoref{figure-DAR}. Moreover, the constraint violations caused by downsampling, \textit{i.e.}, $\textnormal{US}(\Lambda_k)$ is applied to match $x^{k-1}_{s_k^-}$ in \eqref{Parallel-ODE-Training-Task-DS} instead of $\lambda_k$, can be adjusted through the layer-serial training procedure, which also works for applications without the use of data augmentation. Experimental results in \autoref{section-experiments} validate our theoretical findings.


\section{Experiments} \label{section-experiments}

\begin{table*}[t]

\vspace{-0.4cm}
\caption{Memory, test accuracy and SURs of ResNet-110 on CIFAR-10, where $K=2$, $3$ and hybrid ratio $\gamma_H = 1:4$ .}
\vskip 0.05in
\begin{center}
\begin{small}
\begin{tabular}{lccc|lccc}
\toprule
Method & Memory (GB) &Test Acc. & SURs & Method & Memory (GB) & Test Acc. & SURs \\
\midrule
Serial w/o DA & - &  86.4 &  - & Serial with DA & - &  93.7 &  -\\
\hline
Penalty ($K=2$) & 1.53  & 85.0 & 1.41 & AL ($K=2$)& 3.05  & 85.7 & 1.36\\
Penalty ($K=3$)  & 4.58 & 84.5 & 1.56 & AL ($K=3$)      & 9.12 & 85.2 & 1.49\\
\hline
DS-P ($K=2$)  & 0.38 & 84.2 & 1.53 & DS-AL ($K=2$) & 0.57 & 84.1 & 1.42 \\
DS-P ($K=3$)  & 1.44 & 83.7 & 1.69 & DS-AL ($K=3$)  & 2.29 & 83.7 & 1.58 \\
\hline
SPH-P ($K=2$) & 1.53 & 91.8 & 1.30 & SPH-AL ($K=2$)& 3.05 & 91.8 & 1.27 \\
SPH-P ($K=3$) & 4.58 & 91.3 & 1.40 & SPH-AL ($K=3$)& 9.12 & 91.4 & 1.35 \\
\hline
DS-SPH-P ($K=2$) & 0.38 & 91.8 & 1.38 & DS-SPH-AL ($K=2$) & 0.57 & 91.6 & 1.31\\
DS-SPH-P ($K=3$) & 1.44 & 91.8 & 1.48 &DS-SPH-AL ($K=3$) & 2.29 & 91.5 & 1.41\\
\bottomrule
\end{tabular}
\end{small}
\end{center}
\label{table-cifar10}

\vspace{-0.2cm}

\caption{Memory, test accuracy and SURs of WideResNet on CIFAR-100, where $K=3$ and hybrid ratio $\gamma_H = 1:4$.}
\label{table-cifar100}
\vskip 0.05in
\begin{center}
\begin{small}
\begin{tabular}{lccc|lccc}
\toprule
Method & Memory (GB) & Test Acc. & SURs & Method & Memory (GB) & Test Acc. & SURs\\
\midrule
Serial w/o DA & - & 66.53 & - & Serial with DA & - & 80.71 &  - \\
\hline
Penalty & 45.77 & 64.91 & 1.89& AL& 91.55 & 64.80 & 1.67\\
\hline
DS-P & 11.44 & 61.06 & 2.19&  DS-AL & 22.89 & 60.84 & 1.92  \\
\hline
SPH-P & 45.77 & 75.25 & 1.52& SPH-AL & 91.55 & 76.25 & 1.42 \\
\hline
DS-SPH-P & 11.44  & 76.23 & 1.76 & DS-SPH-AL & 22.89  & 76.84 & 1.62  \\
\bottomrule
\end{tabular}
\end{small}
\end{center}
\vskip -0.05in
\end{table*}





To verify our method's effectiveness, we conduct experiments on various settings, including different datasets, network architectures and tasks. Firstly, in Section~\ref{ResNet_cls} we verify our method with image classification on CIFAR-10 and CIFAR-100, respectively. We show that the proposed SPH and DS can effectively deal with the issues on data augmentation and communication, respectively. Moreover, the combination of DS and SPH (DS-SPH) achieves comparable performance with traditional serially training methods while maintaining a competitive speed-up ratio. Secondly, in Section~\ref{ResNet_VAE} we conduct the experiment on image generation to show that our method is not limited to classification. As data augmentation is usually not considered in image generation, our method shows a significantly speed-up ratio (SUR) with quite well generation performance. For simplicity, we use DS-SPH-P and 
DS-SPH-AL to represent penalty based DS-SPH method and AL-based DS-SPH method.

All of our experiments are implemented in Pytorch 1.4. Our model is split into $K$ stages, distributed on $K$ GPUs (Tesla-V100). The parallel methods are implemented based on the Pytorch multiprocessing library with NCCL backends. 


\subsection{Images Classification}\label{ResNet_cls}

In this section, we verify our methods for classification tasks on CIFAR-10 and CIFAR-100. 

\textbf{Implementation Details:} For CIFAR-10, the serial model is implemented based on \textbf{ResNet-110}. For penalty and AL methods, the networks are divided into \textbf{two or three stages}. Empirically, we set the initial penalty coefficient $\beta=100$. For all experiments, we use SGD optimizer with initial learning rate 0.1. We train our models for 200 epochs and decay the learning rate (lr) by 0.1 for every 50 epochs. For CIFAR-100, the serial model is implemented based on \textbf{WideResNet} with 40 layers and widen factor 10 (WideResNet-40-10). We set the penalty coefficient $\beta=10$ as we empirically found it works better for WideResNet. For all experiments on CIFAR-100, we use SGD optimizer with initial learning rate 0.1. We train our models for 200 epochs and decay lr according to cosine-lr schedule. Results of CIFAR-10 and CIFAR-100 are summarized in \autoref{table-cifar10} and \autoref{table-cifar100}.

 \textbf{Benefit of Serial-Parallel-Hybrid Strategy}. The performance of vanilla penalty and AL methods are much worse than the traditional serial training with data augmentation (with approximately 8\% accuracy drop from 93.7\% to 85.7\% for AL $K=2$, see \autoref{table-cifar10}). This is mainly caused by lacking data augmentation (DA), as analyzed in the last section. But with our 1:4 hybrid ratio training strategy, parallel training with 80\% epochs before serial training for the other 20\%, SPH-AL $K=2$ gets test accuracy 91.8\%, reducing the gap to serial training with DA from 8\% to 2\%. This remarkable improvement only causes a slight drop in the speed-up ratio, which is acceptable in practice. Besides, the AL methods are slightly better than the penalty method in terms of test accuracy while at the cost of higher memory load and lower speed-up ratio.
 
\textbf{Benefit of Down-Sampling Strategy.} Apart from the DA problem, another critical problem that limits the penalty and AL parallel methods' efficiency is the storage and CPU-GPU communication of auxiliary variables. Due to the large size of auxiliary variables, these variables cannot be stored directly in GPU memory. We have to save them in the CPU memory and load them to GPU in the training process. The frequent CPU-GPU communication causes a huge communication overhead. Comparing the results of penalty and downsampling with the penalty (DS-P), downsampling the feature maps hurts the performance less than 1\% (decrease from 84.5\% to 83.7\% with $K=3$ in \autoref{table-cifar10}) but significantly reduces the memory cost at least three times ($4.58/1.44\approx3.18$) and increases the speed-up ratio 1.56 to 1.69. Interestingly, although the down-sampling strategy alone decreases the test accuracy slightly, it improves the performance by 0.5\% when combined with SPH. 
 
 \textbf{More Complicate Model and Dataset}. We have shown that our method works well on CIFAR-10 with ResNet. For more complicated datasets and models, our methods still work well. We also verify our method on CIFAR-100 with WideResNet. The feature map of WideResNet is much larger than ResNet. Specifically, the feature map of WideResNet-40-10 is ten times larger than that of ResNet. Therefore, we needs to introduce many auxiliary variables for WideResNet, which makes the parallelization even more challenging. From \autoref{table-cifar100} we see that our method also shows its superiority in training WideResNet. The performance gap between the serial training with data augmentation and vanilla parallel training, which is penalty and AL, is about 15\%, much larger than that of CIFAR-10 with ResNet. Our method decreases this gap to 4.48\% and 3.7\% for penalty and AL, respectively. Furthermore, the speed-up ratio is larger than that in ResNet, say 1.76 compares to 1.48 for DS-SPH-P and 1.62 compares to 1.41 for DS-SPH-AL. It is because a large number of auxiliary variables naturally lead to larger memory cost and communication overhead. Our DS strategy reduces much cost in them and exhibits a more speed-up ratio.

 \textbf{Accuracy-Speed Trade-off}. In this part, we study the accuracy-speed trade-off from two aspects, the number of stages K and the hybrid ratio. Results of $K=2$ and $K=3$ for different methods on CIFAR-10 are shown in \autoref{table-cifar10}. The more stages we divide the model, the more auxiliary variables we need to introduce and the more computing resource we need. Therefore, in general, the memory cost and speed-up ratio grow as K grows large. However, more auxiliary variables also introduce more constraints to the optimization problem making it more difficult to solve. Thus, more stages often leads to a performance drop, while such drop may be negligible. These results are observed in \autoref{table-cifar10} for all methods. More specifically, when increasing K from 2 to 3 DS-SPH-AL's performance drops from 91.6\% to 91.5\% with the speed-up ratio increasing from 1.31 to 1.41. 
 There is no performance difference up to the first decimal when increasing K from 2 to 3 for DS-SPH-P.

 \begin{figure}[t]
\begin{center}
\centerline{\includegraphics[width=0.75\columnwidth]{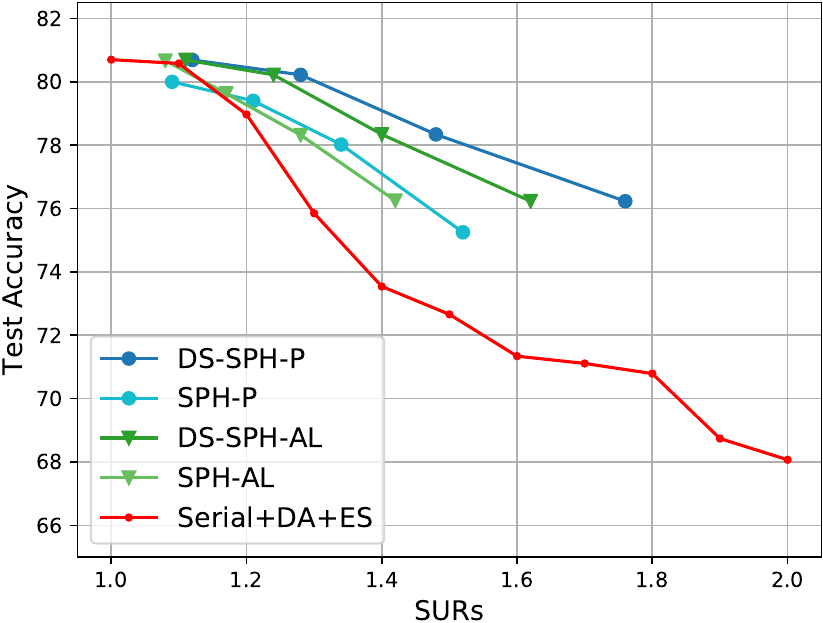}}
\caption{Accuracy-speed trade-off for different methods. The red-dot line is drawn by early stopping (ES) the serial training at epochs corresponding to the speed-up ratio. The DS-SPH methods dominate other methods.}
\vspace{-0.2cm}
\label{figure-SPP-ES-cifar100}
\end{center}
\vspace{-0.8cm}
\end{figure}

 The other factor which affects the accuracy-speed trade-off is the hybrid ratio $\gamma_H$. In \autoref{table-cifar10-DS-SPH} and \autoref{table-cifar100-DS-SPH}, we show the results of different $\gamma_H$ on CIFAR-10 and CIFAR-100, respectively. As $\gamma_H$ increases, the speed-up ratio decreases accordingly, but the performances become closer to the serial training results. To be more specific, take DS-SPH-P as an example, when $\gamma_H = \text{1:4}$, the test accuracy is 91.25\% with a speed-up ratio of 1.40. As increasing $\gamma_H$ to 4:1, the test accuracy approach to 93.62\%, which is very close to the serial training result of 93.70\%. In \autoref{figure-SPP-ES-cifar100}, we compare this trade-off of CIFAR-100 between different methods. Our DS-SPH method consistently outperforms other methods.
 
\begin{table}[t]
\vspace{-0.3cm}
\caption{For ResNet-110 on CIFAR-10, $\gamma_H$, SURs, test accuracy of SPH and DS-SPH using penalty and AL methods.}
\label{table-cifar10-DS-SPH}
\vskip 0.05in
\begin{center}
\begin{small}
\begin{tabular}{|c|c|cc|cc|}
\toprule
\multirow{2}{*}{$\gamma_H$} & \multirow{2}{*}{Methods} & \multicolumn{2}{c|}{Penalty} & \multicolumn{2}{c|}{AL}  \\
\cline{3-6} 
 &  & SURs & Test Acc.  & SURs & Test Acc. \\
\midrule
1:4 & SPH & 1.40 & 91.25   & 1.35 & 91.38 \\
2:3 & SPH &  1.27 & 92.36  & 1.24 & 92.64 \\
3:2 & SPH & 1.16 & 92.85  & 1.14 & 93.13 \\
4:1 & SPH & 1.07 & 93.18   & 1.07 & 93.75  \\
\hline
1:4 & DS-SPH &  1.48 & 91.75  & 1.41 & 91.54 \\
2:3 & DS-SPH &  1.32 & 92.79  & 1.28 & 92.58 \\
3:2 & DS-SPH &  1.19 & 93.58  & 1.17 & 93.11 \\
4:1 & DS-SPH &  1.08 & 93.62 & 1.08 & 93.66  \\
\hline
- & Serial+DA & - & 93.70 & ~ &  ~ \\
\bottomrule

\end{tabular}
\end{small}
\end{center}

\caption{For WideResNet on CIFAR-100, $\gamma_H$, SURs, test accuracy of SPH and DS-SPH using penalty and AL methods.}
\label{table-cifar100-DS-SPH}
\vskip 0.05in
\begin{center}
\begin{small}
\begin{tabular}{|c|c|cc|cc|}
\toprule
\multirow{2}{*}{$\gamma_H$} & \multirow{2}{*}{Methods} & \multicolumn{2}{c|}{Penalty} & \multicolumn{2}{c|}{AL}  \\
\cline{3-6} 
 & & SURs & Test Acc.  & SURs & Test Acc. \\
\midrule
1:4 & SPH &  1.52 & 75.25  & 1.42 & 76.25 \\
2:3 & SPH  & 1.34 & 78.02  & 1.28 & 78.32\\
3:2 & SPH & 1.21 & 79.40  & 1.17 & 79.63 \\
4:1 & SPH & 1.09 & 80.00  & 1.08 & 80.67  \\
\hline
1:4 & DS-SPH & 1.76 & 76.23  & 1.62 & 76.84 \\
2:3 & DS-SPH & 1.48 & 78.34  & 1.40 & 78.08 \\
3:2 & DS-SPH & 1.28 & 80.22  & 1.24 & 79.97 \\
4:1 & DS-SPH &  1.12 & 80.69 & 1.11 & 80.37  \\
\hline 
-   & Serial+DA & - & 80.70 &  ~ & ~\\
\bottomrule

\end{tabular}
\end{small}
\end{center}
\vspace{-0.2cm}
\end{table}

\subsection{Image Generation} \label{ResNet_VAE}

This section shows the experimental results on the image generation task to demonstrate that our method is not limited to image classification.

\textbf{Implementation Details:} A VAE \cite{kingma2013auto} model contains a pair of encoder and decoder. The serial model is implemented based on \textbf{ResNet-VAE} whose encoder is \textbf{ResNet-110}. The encoder for penalty method and AL method is divided to \textbf{three stages} and the initial penalty coefficient $\beta$ is set as 10. We use Adam as an optimizer for all networks, and the initial learning rate is set as 0.01. To evaluate the generated image, we use the reconstructing \textbf{MSE loss} between reconstructed images and original images on the test set.

\begin{figure}[t]
\begin{center}
\centerline{\includegraphics[width=0.7\columnwidth]{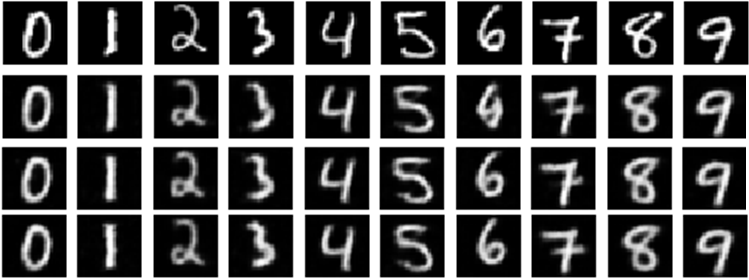}}
\caption{Results for reconstruction images in test set. \textbf{Line-1}: Original images; \textbf{Line-2}: Reconstruction from serial method; \textbf{Line-3}: Reconstruction from DS-SPH using penalty method; \textbf{Line-4}: Reconstruction images from DS-SPH using AL method.}
\label{figure-mnist-reconstruction}
\end{center}
\vspace{-0.5cm}
\end{figure}

The results are shown in \autoref{table-mnist}. As data augmentation is usually not used in the training of VAE, we omit the experiments with only SPH. As the results show, the vanilla penalty and AL, in general, can achieve acceptable performance on VAE, since data augmentation is not essential. However, they still suffer from enormous memory cost and CPU-GPU communication overhead. DS strategy reduces those costs significantly with a slight drop in performance. The reconstruction loss reduces from 0.149 to 0.160 for DS-P and from 0.129 to 0.137 for DS-AL. Combining with SPH, the performance drop caused by DS is almost eliminated. This result shows that SPH plays a more significant role than merely providing data augmentation. It also helps eliminate the gap between the real feature map and the auxiliary feature map and get better results, in \autoref{figure-mnist-reconstruction} we show the reconstructed images sampled from the test set to examine their visual qualities. From the figure, we see that our methods indeed reconstruct high-quality images.

\begin{table}[t]
\vspace{-0.3cm}
\caption{Test loss and SURs based on above two methods with K=3 and hybrid ratio $\gamma_H =\text{1:4}$ on MNIST Generation.}
\label{table-mnist}
\vskip 0.15in
\begin{center}
\begin{small}
\begin{sc}
\begin{tabular}{lcc}
\toprule
Method & Test Loss & SUR\\
\midrule
serial model & 0.126 & - \\
Penalty         & 0.149 & 2.33 \\
AL              & 0.129  & 2.18 \\
\hline
DS-P & 0.160 & 2.41\\
DS-AL & 0.137 & 2.25\\
\hline
DS-SPH-P & 0.127 & 1.88\\
DS-SPH-AL & 0.128 & 1.80\\
\bottomrule
\end{tabular}
\end{sc}
\end{small}
\end{center}
\end{table}

\section{Conclusion}

In this paper, we observed that the key issues that hampered the practicality of layer-parallel training were data augmentation and communication. We then proposed a novel hybrid training strategy combined with downsampling to resolve the aforementioned issues, and demonstrated the effectiveness of the proposed method on training large residual networks on CIFAR-10 and CIFAR-100. Potential future directions include investigation on the proposed method with more heavy duty deep residual networks, larger number of stages, exploring other choices of downsampling operators, other layer-parallel training algorithms, etc.



\newpage

\nocite{langley00}

\bibliography{references}
\bibliographystyle{icml2021}

\newpage
\onecolumn

\centerline{\Large{\textbf{Supplementary Material}}}

\appendix

\section{Layer-Serial Training of Residual Networks} \label{appendix-ResNet}

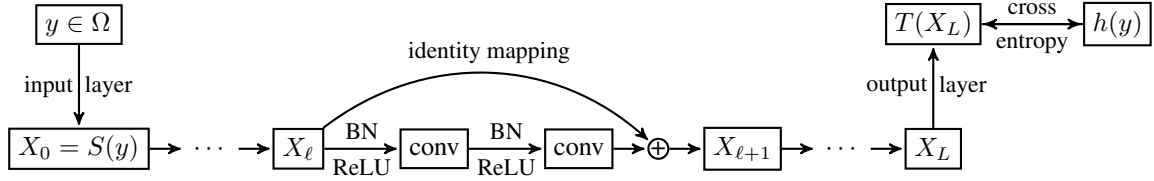
\begin{figure}[h]
\begin{adjustbox}{max totalsize={\textwidth}{\textheight},center}
\begin{tikzpicture}[shorten >=1pt,auto,node distance=2.5cm,thick,main node/.style={rectangle,draw,font=\normalsize},decoration={brace,mirror,amplitude=7}]
\node[main node] (1) {$X_\ell$};
\node[main node] (2) [right = 1cm of 1] {\parbox[c][0.28cm]{0.66cm}{conv} };
\node[main node] (3) [right = 1cm of 2] {\parbox[c][0.28cm]{0.66cm}{conv} };
\node[draw, circle, inner sep=0pt, minimum size=0.21cm, color=black] (4) [right = 0.45cm of 3] {+};
\node[main node] (5) [right = 0.45cm of 4]  {$X_{\ell+1}$};

\node[] (6) [left = 0.45cm of 1]  {$\cdots$};
\node[main node] (7) [left = 0.45cm of 6]  {\parbox[c][0.3cm]{1.62cm}{$X_0=S(y)$} };
\node[] (8) [right = 0.45cm of 5]  {$\cdots$};
\node[main node] (9) [right = 0.45cm of 8]  {$X_{L}$};

\node[main node] (10) [above = 1.1cm of 7] {\parbox[c][0.28cm]{0.9cm}{$y\in \Omega$} };
\node[main node] (11) [above = 1.1cm of 9] {\parbox[c][0.3cm]{1.02cm}{$T(X_L)$} };
\node[main node] (12) [right = 1.35cm of 11] {\parbox[c][0.33cm]{0.65cm}{$h(y)$} };

\path[every node/.style={font=\sffamily\small},->,>=stealth']
    (1) edge node[above,midway] {\textnormal{BN}} node[below,midway] {\textnormal{ReLU}} (2)
    		edge [bend left=40] node[above] {\textnormal{identity mapping}} (4)
    (2) edge node[above,midway] {\textnormal{BN}} node[below,midway] {\textnormal{ReLU}} (3)
    (3) edge node {} (4)
    (4) edge node {} (5)
    (7) edge node {} (6)
    (6) edge node {} (1)
    (5) edge node {} (8)
    (8) edge node {} (9);

\path[every node/.style={font=\sffamily\small},->,>=stealth']
    (10) edge node[left] {\textnormal{input}\!} node[right] {\!\textnormal{layer}} (7)
    (9) edge node[left] {\textnormal{output}\!} node[right] {\!\textnormal{layer}} (11);
\path[every node/.style={font=\sffamily\small},->,>=stealth']    
    (11) edge node[above,midway] {\textnormal{cross}} node[below,midway] {\textnormal{entropy}} (12)
    (12) edge (11);  
\end{tikzpicture}
\end{adjustbox}
\vspace{-0.5cm}
\caption{A diagram describing the serial training process of a pre-activation ResNet.}
\label{fig-ResNet-architecture}
\end{figure}

Without loss of generality, we consider the benchmark residual learning framework \cite{he2016deep,he2016identity} that assigns pixels in the raw input image to categories of interest as depicted in \autoref{fig-ResNet-architecture}. Its continuous-time analogue \cite{thorpe2018deep,weinan2017proposal} is then introduced to bridge such an image classification task with a terminal control problem constrained by the so-called neural ordinary differential equation \cite{chen2018neural}.

Given a human-labeled database $\{y,h(y)\}_{y\in\Omega}$, the optimization of network parameters requires solving problem \eqref{ResNet-Training-Task}, \textit{i.e.},
\begin{equation*}
	\operatorname*{arg\,min}_{\{W_\ell\}_{\ell=0}^{L-1}} \left\{ \mathbb{E}_{y\in \Omega} \! \Big[ \lVert T(X_L) - h(y) \rVert \Big] \, \Big| \,  X_0=S(y), \ X_{\ell+1} = X_\ell + F(X_\ell,W_\ell) \ \ \text{for}\ 0\leq \ell\leq L-1 \right\} 
\end{equation*}
where $X_\ell$ indicates the input feature map of the $\ell$-th building module, $L\in\mathbb{N}_+$ the total number of modules, $F$ typically a composition of linear and nonlinear functions as depicted in \autoref{fig-ResNet-architecture}, $W_\ell$ the network parameters to be learned, and $\lVert\cdot\rVert$ a given metric measuring the discrepancy between the model prediction $T(X_L)$ and the ground-truth label $h(y)$ for each training sample $y\in\Omega$. The trainable parameters of input and output layers, \textit{i.e.}, $S$ and $T$ in \autoref{fig-ResNet-architecture}, are assumed to be fixed \cite{haber2018learning} for the ease of illustration.

When the most commonly used backpropagation algorithm \cite{hecht1992theory} is applied to solving the optimization problem \eqref{ResNet-Training-Task}, we obtain formula \eqref{ResNet-BackPropagation} for parameter updates, \textit{i.e.}, 
\begin{equation*}
	W_\ell \leftarrow W_\ell - \eta \frac{\partial \varphi(X_L)} {\partial W_\ell} = W_\ell - \eta \frac{\partial \varphi(X_L)}{\partial X_{\ell+1}}\frac{\partial X_{\ell+1}}{\partial W_\ell}, \qquad 0\leq \ell\leq L-1,
\end{equation*}
where $\eta>0$ is the learning rate and $\displaystyle \varphi(X_L)=\mathbb{E}_{y\in \Omega} \! \Big[ \lVert T(X_L) - h(y) \rVert \Big]$ \footnote{Though the population risk is of primary interest, we only have access to the empirical risk in practice. For notational simplicity, we still denote by $\varphi(\cdot)$ the objective function obtained from a mini-batch of the entire training data throughout this work.}.

Note that by defining $\displaystyle P_{\ell+1} = \frac{\partial \varphi(X_L)}{\partial X_{\ell+1}}$ for $0\leq\ell\leq L-1$, formula \eqref{ResNet-BackPropagation} can be rewritten as
\begin{equation}
	W_\ell \leftarrow W_\ell - \eta \left( P_{\ell+1}\frac{\partial F(X_\ell,W_\ell)}{\partial W}\right), \qquad 0\leq \ell \leq L-1,
	\label{ResNet-Parameter-Updates}
\end{equation}
where ${\{P_{\ell+1}\}_{\ell=0}^{L-1}}$ satisfy a backward dynamic that captures the objective changes with respect to hidden neurons, \textit{i.e.},
\begin{equation}
	P_\ell = P_{\ell+1}\frac{\partial X_{\ell+1}}{\partial X_\ell}  = P_{\ell+1} + P_{\ell+1} \frac{\partial F(X_\ell,W_\ell)}{\partial X}, \qquad  P_L = \frac{\partial \varphi(X_L)}{\partial X_L}.
	\label{ResNet-Gradient-Propagation}
\end{equation}
To put it differently, the full serial backpropagation algorithm \eqref{ResNet-BackPropagation} is handled by formulae \eqref{ResNet-Gradient-Propagation} and \eqref{ResNet-Parameter-Updates}. Therefore, the training of ResNets at each iteration step requires the repeated execution of
\begin{equation*}
	\bullet \ \ \textnormal{forward pass in \eqref{ResNet-Training-Task}} \qquad \bullet \ \ \textnormal{backward gradient propagation}\ \eqref{ResNet-Gradient-Propagation} \qquad \bullet \ \ \textnormal{parameter updates}\ \eqref{ResNet-Parameter-Updates}
\end{equation*}
which can be very time-consuming as it is common to see neural networks with hundreds or even thousands of layers.

\subsection{Optimal Control of Neural Ordinary Differential Equations}

The continuous-time counterpart of the minimization problem \eqref{ResNet-Training-Task} is formulated as \eqref{ODE-Training-Task}, that is,
\begin{equation*}
	\operatorname*{arg\,min}_{\omega_t} \left\{ \mathbb{E}_{y\in \Omega} \! \Big[ \lVert T(x_1) - h(y) \rVert \Big] \, \Big| \, x_0=S(y),\ dx_t = f(x_t,w_t)dt \ \ \text{for}\ 0< t\leq 1 \right\}
\end{equation*}
where the forward propagation through the underlying network with fixed parameters, \textit{i.e.}, the constraint of \eqref{ResNet-Training-Task} is interpreted as a numerical discretization of differerntial equations \cite{weinan2017proposal,lu2017beyond,chen2018neural}.

By introducing the Lagrange functional with multiplier $p_t$ \cite{nocedal2006numerical}, solving the constrained optimization problem \eqref{ODE-Training-Task} is equivalent to finding saddle points of the following Lagrange functional without constraints\footnote{For notational simplicity, $\frac{dx_t}{dt}$ and $\dot{x}_t$ are used to denote the time derivative of $x_t$ throughout this work.}
\begin{equation*}
\begin{split}
	\mathcal{L}(x_t,w_t,p_t) & = \varphi(x_1) + \int_0^1 p_t\left(f(x_t,w_t)-\dot{x}_t\right) dt \\
	& =  \varphi(x_1) - p_1x_1 + p_0x_0 + \int_0^1 p_tf(x_t,w_t) + \dot{p}_tx_t\,dt.
\end{split}
\end{equation*}
and the variation in $\mathcal{L}(x_t,w_t,p_t)$ corresponding to a variation $\delta w$ in control $w$ takes on the form \cite{liberzon2011calculus}
\begin{equation*}
	\delta\mathcal{L} = \bigg[ \frac{\partial \varphi(x_1)}{\partial x} - p_1 \bigg] \delta x + \int_0^1 \bigg( p_t\frac{\partial f(x_t,w_t)}{\partial x} + \dot{p}_t \bigg)\delta x + \bigg( p_t\frac{\partial f(x_t,w_t)}{\partial w} \bigg) \delta w \, dt,
\end{equation*}
which leads to the necessary conditions for $w_t=w_t^*$ to be the extremal of $\mathcal{L}(x_t,w_t,p_t)$, \textit{i.e.},
\begin{subequations}
  \begin{align*}
    & dx_t^* = f(x_t^*,w_t^*)dt, & & x_0^*=S(y), & & (\textnormal{state equation}) \\
    & dp_t^* = -p_t^* \frac{\partial f(x_t^*,w_t^*)}{\partial x}dt, & & p_1^*=\frac{\partial \varphi(x_1^*)}{\partial x}, & &  (\textnormal{adjoint equation}) \\
    & p_t^* \frac{\partial f(x_t^*,w_t^*)}{\partial w} = 0, & & 0\leq t\leq 1. & & (\textnormal{optimality condition})  
  \end{align*}
\end{subequations}
However, directly solving this optimality system is computationally infeasible due to the so-called curse of dimensionality, a gradient-based iterative approach with step size $\eta>0$ is typically used, \textit{e.g.},
\begin{subequations}\label{ResNet-ODE-KKT-System}
  \begin{align}
    & dx_t = f(x_t,w_t)dt, & & x_0=S(y), & & (\textnormal{forward pass})  \\ 
    & dp_t = -p_t \frac{\partial f(x_t,w_t)}{\partial x}dt, & & p_1=\frac{\partial \varphi(x_1)}{\partial x}, & &  (\textnormal{backward gradient propagation})  \\ 
	& w_t \leftarrow w_t - \eta \left( p_t \frac{\partial f(x_t,w_t)}{\partial w} \right), & &  0\leq t\leq 1, & &  (\textnormal{parameter updates}) 
  \end{align}
\end{subequations}
which is consistent with the layer-serial training of ResNet through forward-backward propagation, \textit{i.e.}, \eqref{ResNet-Training-Task}, \eqref{ResNet-Gradient-Propagation} and \eqref{ResNet-Parameter-Updates}, by taking the limit as $L\to\infty$ \cite{li2017maximum}. In other words, the backprapagation approach \eqref{ResNet-BackPropagation} can be recovered from \eqref{ODE-Adjoint-Equation} and \eqref{ODE-Weights-Update} by employing the stable discretization schemes \eqref{ResNet-Gradient-Propagation} and \eqref{ResNet-Parameter-Updates}.

\section{Augmented Lagrangian Method} \label{appendix-Calculsu-Variations}

Recall that the neural ODE-constrained optimization problem \eqref{ODE-Training-Task} can be reformulated as \eqref{ODE-Training-Task-ReWritten}, that is,
\begin{equation*}
	\operatorname*{arg\,min}_{\{w_t^k\}_{k=0}^{K-1}} \left\{ \varphi(x^{K-1}_{s_K^-}) \, \Big| \, x^{k-1}_{s_{k}^-} = \lambda_k \ \ \textnormal{and} \ \ x^k_{s_k^+}=\lambda_k, \   d x^k_t = f(x^k_t,w^k_t)dt\ \ \textnormal{on}\ (s_k,s_{k+1}] \ \ \textnormal{for}\ 0\leq k \leq K-1 \right\},
\end{equation*}
whose the augmented Lagrangian functional is expressed as
\begin{equation*}
\begin{split}
	& \mathcal{L}_{\textnormal{AL}}( x_t^k, p_t^k, w^k_t, \lambda_k,\kappa_k ) = \varphi(x^{K-1}_{s_K^-})  + \sum_{k=0}^{K-1} \left( \beta \psi(\lambda_k,x^{k-1}_{s_{k}^-}) - \kappa_k( \lambda_k - x^{k-1}_{s_{k}^-} ) + \int_{s_k}^{s_{k+1}} p^k_t \big( f(x_t^k,w^k_t) - \dot{x}_t^k \big)dt \right) \\
 	= \ & \varphi(x^{K-1}_{s_K^-})  + \sum_{k=0}^{K-1} \left( \beta \psi(\lambda_k,x^{k-1}_{s_{k}^-}) - \kappa_k( \lambda_k - x^{k-1}_{s_{k}^-} ) - p^k_{s_{k+1}^-}x^k_{s_{k+1}^-} + p^k_{s_{k}^+}\lambda_k +  \int_{s_k}^{s_{k+1}} p^k_t f(x_t^k,w^k_t) + \dot{p}_t^k x_t^k\,dt \right).
\end{split}	
\end{equation*}

Specifically, the augmented Lagrangian functional can be decomposed as parts involving $x_t^{K-1}$ and $\{x_t^k\}_{k=0}^{K-2}$, \textit{i.e.},
\begin{equation*}
		\RNum{1} = \varphi(x^{K-1}_{s_K^-}) - p_{s_K^-}^{K-1}x_{s_K^-}^{K-1} + p_{s_{K-1}^+}^{K-1}\lambda_{K-1} + \int_{s_{K-1}}^{s_K} p_t^{K-1}f(x_t^{K-1},w^{K-1}_t) + \dot{p}_t^{K-1}x_t^{K-1}\,dt,
\end{equation*}
and 
\begin{equation*}
		\RNum{2} = \sum_{k=0}^{K-2} \left( \beta \psi(\lambda_{k+1},x^k_{s_{k+1}^-}) + (\kappa_{k+1} - p^k_{s_{k+1}^-})x^k_{s_{k+1}^-} + p^k_{s_{k}^+}\lambda_k + \int_{s_k}^{s_{k+1}} p_t^{k}f(x_t^{k},w^k_t) + \dot{p}_t^{k}x_t^{k}\,dt -\kappa_{k+1}\lambda_{k+1} \right)
\end{equation*}
respectively, then the variation in $\mathcal{L}( x_t^k, p_t^k, w^k_t, \lambda_k,\kappa_k )$ corresponding to a variation $\delta w_t^k$ in control $w^k_t$ takes on the form
\begin{equation*}
\begin{split}
	\delta\mathcal{L} = & \left( \frac{\partial \varphi(x^{K-1}_{s_K^-})}{\partial x} - p_{s_K^-}^{K-1} \right) \delta x^{K-1} + \int_{s_{K-1}}^{s_K} \left( p_t^{K-1}\frac{\partial f(x_t^{K-1},w^{K-1}_t)}{\partial x} + \dot{p}_t^{K-1} \right)\delta x^{K-1} \, dt \\
	& + \sum_{k=0}^{K-2} \left[ \left( \beta \frac{\partial \psi(\lambda_{k+1},x^k_{s_{k+1}^-})}{\partial x} + \kappa_{k+1} - p_{s_{k+1}^-}^k \right) \delta x^k + \int_{s_k}^{s_{k+1}} \left( p_t^k\frac{\partial f(x_t^k,w^k_t)}{\partial x} + \dot{p}_t^k \right)\delta x^k \, dt \right],
\end{split}	
\end{equation*} 
which implies that the adjoint variable $p_t^k$ satisfies the backward differential equations \eqref{Parallel-ODE-Adjoint-Equation-AL} \cite{liberzon2011calculus}, namely,
\begingroup
\renewcommand*{\arraystretch}{3}
\vspace{-0.2cm}
\begin{equation}
\begin{array}{l}
	\displaystyle dp_t^{k} = - p_t^{k}\frac{\partial f(x_t^{k},w^k_t)}{\partial x} dt \ \ \textnormal{on}\ [s_k,s_{k+1}),\\
	\displaystyle p^k_{s_{k+1}^-} = \left(1-\delta\right) \left( \beta \frac{\partial \psi(\lambda_{k+1},x^k_{s_{k+1}^-})}{\partial x} + \kappa_{k+1} \right) +\delta \, \frac{\partial \varphi(x^{k}_{s_{k+1}^-})}{\partial x},	
\end{array}
\label{Parallel-ODE-Adjoint-Equation-AL}
\end{equation}
\endgroup
for any $0\leq k\leq K-1$. Here and in what follows $\delta=\delta_{k,K-1}$ represents the Kronecker Delta function. 

Moreover, it can be easily deduced from the augmented Lagrangian functional that the control updates satisfy
\begin{equation}
	w_t^k \leftarrow w_t^k - \eta \left( p_t^{k} \frac{\partial f(x_t^k,w^k_t)}{\partial w}\right) \ \ \ \ \textnormal{on} \ \ [s_k,s_{k+1}]
	\label{Parallel-ODE-Weights-Update-AL}
\end{equation}
for $0\leq k\leq K-1$, while the correction of auxiliary variables takes on the form
\begin{equation}
	\lambda_0 \equiv x_0 \qquad \textnormal{and} \qquad \lambda_k \leftarrow \lambda_k - \eta \Bigg( \beta \frac{\partial \psi(\lambda_k,x^{k-1}_{s_{k}^-})}{\partial \lambda} + p^k_{s_k^+} - \kappa_k \Bigg) \ \ \textnormal{for} \ \, 1\leq k\leq K-1.
	\label{Parallel-ODE-Slack-Variables-Update-AL}
\end{equation}
Notably, by choosing a quadratic penalty function $\psi(\lambda,x) = \lVert \lambda-x \rVert_{\ell_2}^2$, formula \eqref{Parallel-ODE-Slack-Variables-Update-AL} shows that the constraint violations associated with the minimizer of augmented Lagrangian method satisfy for $1\leq k\leq K-1$,
\begin{equation}
	  \lambda_k - x^{k-1}_{s_{k}^-}  \approx \frac{1}{2\beta} (\kappa_k - p^k_{s_k^+})
	  \label{Constraint-Violation-AL}
\end{equation}
which offers two ways of improving the consistency constraint $x^{k-1}_{s_{k}^-}=x^{k}_{s_{k}^+}$: increasing $\beta$ or sending $\kappa_k\to p^k_{s_k^+}$, whereas the penalty method (by forcing $\kappa_k\equiv 0$ in \eqref{Constraint-Violation-AL}, see also the formula \eqref{Constraint-Violation-Penalty} below) provides only one option. Moreover, it can be deduced from \eqref{Constraint-Violation-AL} that the update rule of explicit Lagrange multipliers satisfy
\begin{equation}
	\kappa_0 \equiv 0 \qquad \textnormal{and} \qquad \kappa_k \leftarrow \kappa_k - \frac{\eta}{2\beta} \left(\lambda_k -x^{k-1}_{s_k^-}\right)\ \ \textnormal{for} \ \, 1\leq k\leq K-1.
	\label{Parallel-ODE-Multiplier-Update-AL}
\end{equation}

In short, the augmented Lagrangian method for approximately solving problem \eqref{ODE-Training-Task} at each iteration step includes 
\begin{tcolorbox}[
						  standard jigsaw,
    						  opacityback=0,  
						]
\begin{empheq}{align*}
	\bullet \ \textnormal{local operations}\ \eqref{Parallel-ODE-State-Equation}, \eqref{Parallel-ODE-Adjoint-Equation-AL}, \eqref{Parallel-ODE-Weights-Update-AL}\ \textnormal{in parallel} \ \qquad \ \bullet \ \textnormal{global communication}\ \eqref{Parallel-ODE-Slack-Variables-Update-AL}, \eqref{Parallel-ODE-Multiplier-Update-AL}
\end{empheq}
\end{tcolorbox}
which not only parallelizes the iterative system \eqref{ResNet-ODE-KKT-System} for solving \eqref{ODE-Training-Task} but also lessens the the issue of coefficient tuning.

\subsection{Penalty Method}
Note that by forcing $\kappa_k\equiv 0$ for any $0\leq k\leq K-1$, the augmented Lagrangian method degenerates to a penalty method. Specifically, it can be deduced from \eqref{Parallel-ODE-Adjoint-Equation-AL} that the adjoint equation for relaxed minimization problem \eqref{Parallel-ODE-Training-Task} takes on the form
\begingroup
\renewcommand*{\arraystretch}{3}
\vspace{-0.2cm}
\begin{equation}
\begin{array}{l}
	\displaystyle dp_t^{k} = - p_t^{k}\frac{\partial f(x_t^{k},w^k_t)}{\partial x} dt \ \ \ \textnormal{on}\ [s_k,s_{k+1}),\\
	\displaystyle p^k_{s_{k+1}^-} = \left(1-\delta\right) \beta \frac{\partial \psi(\lambda_{k+1},x^k_{s_{k+1}^-})}{\partial x} +\delta \, \frac{\partial \varphi(x^{k}_{s_{k+1}^-})}{\partial x},
\end{array}
	\label{Parallel-ODE-Adjoint-Equation}
\end{equation}
\endgroup
for $0\leq k\leq K-1$. Moreover, by \eqref{Parallel-ODE-Weights-Update-AL} and \eqref{Parallel-ODE-Slack-Variables-Update-AL} , the update rule for control variables now satisfies for $0\leq k\leq K-1$,
\begin{equation}
	w_t^k \leftarrow w_t^k - \eta \left( p_t^{k} \frac{\partial f(x_t^k,w^k_t)}{\partial w}\right) \ \ \ \textnormal{on} \ [s_k,s_{k+1}],
	\label{Parallel-ODE-Weights-Update}
\end{equation}
while the correction of auxiliary variables is given by
\begin{equation}
	\lambda_0 \equiv x_0\ \qquad \textnormal{and}\ \qquad \lambda_k \leftarrow \lambda_k - \eta \Bigg( \beta \frac{\partial \psi(\lambda_k,x^{k-1}_{s_{k}^-})}{\partial \lambda} + p^k_{s_k^+} \Bigg)\ \ \textnormal{for}\ \, 1\leq k\leq K-1.
	\label{Parallel-ODE-Slack-Variables-Update}
\end{equation}

In short, the penalty approach \eqref{Parallel-ODE-Training-Task} for approximately solving problem \eqref{ODE-Training-Task} at each iteration consists of 
\begin{tcolorbox}[
						  standard jigsaw,
    						  opacityback=0,  
						]
\begin{empheq}{align*}
	\bullet \ \textnormal{local operations}\ \eqref{Parallel-ODE-State-Equation}, \eqref{Parallel-ODE-Adjoint-Equation}, \eqref{Parallel-ODE-Weights-Update}\ \textnormal{in parallel} \ \qquad \ \bullet \ \textnormal{global communication}\ \eqref{Parallel-ODE-Slack-Variables-Update}
\end{empheq}
\end{tcolorbox}
which parallelizes the iterative system \eqref{ResNet-ODE-KKT-System} for solving \eqref{ODE-Training-Task}.

In particular, by choosing the quadratic penalty function $\psi(\lambda,x) = \lVert \lambda-x \rVert_{\ell_2}^2$ as before, it can be deduced from \eqref{Parallel-ODE-Slack-Variables-Update} that the constraint violations associated with the approximate minimizer of problem \eqref{Parallel-ODE-Training-Task} satisfies for $1\leq k\leq K-1$,
\begin{equation}
	  \lambda_k - x^{k-1}_{s_{k}^-}  \approx - \frac{1}{2\beta}p^k_{s_k^+}
	  \label{Constraint-Violation-Penalty}
\end{equation}
which implies that a large penalty coefficient $\beta$ is needed in order to force the minimizer of \eqref{Parallel-ODE-Training-Task} close to the feasible region of problem \eqref{ODE-Training-Task}. By employing the augmented Lagrangian method \eqref{Constraint-Violation-AL}, the ill-conditioning of penalty method can be lessened without increasing the penalty coefficient indefinitely, however, the introduction of external Lagrangian multipliers $\{\kappa_k\}_{k=0}^{K-1}$ requires additional memory and communication overheads that may hamper the speed-up ratio. 

\section{Parallel Backpropagation and Communication} \label{appendix-parallel-training-algorithm}

By utilizing the consistent finite difference schemes (see \autoref{appendix-ResNet}) for the discretization of the time-parallel iterative systems established in \autoref{section-parareal-terminal-control}, we arrive at a layer-parallel training algorithm that enables us to fully leveraging the computing resources. The detailed derivations are presented in what follows.

Recall the partitioning of $[0,1]$ associated with the original ResNet \eqref{ResNet-Training-Task}, \textit{i.e.},
\begin{equation*}
	0 = t_0 < t_1 < \ldots < t_\ell = \ell\Delta t < t_{\ell+1} < \ldots < t_{L=nK}=1,
\end{equation*}
then the local sub-problem is built by choosing a coarsening factor $n>1$ and extracting every $n$-th module as depicted in \autoref{fig-multifidelity-forward-propagation}, or, equivalently, the forward Euler discretization of neural ODE with the coarser gird introduced in \autoref{section-parareal-terminal-control}
\begin{equation*}
	t_0 = s_0  < \ldots < s_k = t_{nk} < s_{k+1} < \ldots < s_{K}=t_L,
\end{equation*}
which can be implemented independently and trained with low accuracy at a correspondingly low cost\footnote{The trainable parameters in the input and output layers, \textit{i.e.}, $S$ and $T$, can be automatically updated by coupling into the first and last sub-problems respectively.}. 

To be specific, $[s_k,s_{k+1}]$ is uniformly divided into $n$ sub-intervals for $0\leq k\leq K-1$, \textit{i.e.}, 
\begin{equation*}
	s_k = t_{kn} < t_{kn+1} < \cdots < t_{kn+n-1} < t_{kn+n} = s_{k+1},
\end{equation*}
we have by \eqref{Parallel-ODE-State-Equation} that feature flow of the $k$-th sub-network evolves according to
\begin{equation}
	X_{kn}^k =\lambda_k, \qquad  X_{kn+m+1}^k = X_{kn+m}^k + F(X_{kn+m}^k,W^k_{kn+m})
	\label{Parallel-ResNet-Feature-Flow}
\end{equation}
where $0\leq m\leq n-1$. Then by using the particular numerical scheme \eqref{ResNet-Gradient-Propagation} that arises from the discrete-to-continuum transition in \autoref{section-ResNet}, the discretization of the adjoint equation \eqref{Parallel-ODE-Adjoint-Equation-AL} is given by the backward dynamic
\begingroup
\renewcommand*{\arraystretch}{3}
\vspace{-0.18cm}
\begin{equation}
\begin{array}{l}
	\displaystyle P_{kn+m}^k = P_{kn+m+1}^k  + P_{kn+m+1}^k \frac{\partial F(X_{kn+m}^k,W^k_{kn+m})}{\partial X} = P_{kn+m+1}^k \frac{\partial X^k_{kn+m+1}}{\partial X^k_{kn+m}},\\
	\displaystyle P_{kn+n}^k = (1-\delta) \left( \beta\frac{\partial \psi(\lambda_{k+1},X^k_{kn+n})}{\partial X} + \kappa_{k+1} \right) + \delta  \frac{\partial \varphi(X_{kn+n}^{k})}{\partial X}.
\end{array}
	\label{Parallel-ResNet-Gradient-Propagation-Equation}
\end{equation}
\endgroup

In other words, for any interval $[s_k,s_{k+1}]$ and arbitrary $0\leq m\leq n$,  the adjoint variable in \eqref{Parallel-ResNet-Gradient-Propagation-Equation} is equivalent to
\begin{equation}
	P_{kn+m}^k = (1-\delta) \left( \beta \frac{\partial \psi(\lambda_{k+1},X^k_{kn+n})}{\partial X^k_{kn+m}} + \kappa_{k+1} \frac{\partial X^k_{kn+n}}{\partial X^k_{kn+m}} \right) + \delta \frac{\partial \varphi(X_{kn+n}^{k})}{\partial X^{k}_{kn+m}}
	\label{Parallel-ResNet-Gradient-Propagation}
\end{equation}
which captures the objective and layer-wise synthetic loss changes, namely, the second and the first term on the right-hand-side of \eqref{Parallel-ResNet-Gradient-Propagation}, with respect to the latent states for $k=K-1$ and $0\leq k\leq K-2$, respectively.

Contrary to the straightforward approach \cite{maday2002parareal,gunther2020layer} where the iterations are executed by first solving state equation \eqref{Parallel-ODE-State-Equation}, then adjoint equation \eqref{Parallel-ODE-Adjoint-Equation} afterwards, and finally control updates \eqref{Parallel-ODE-Weights-Update}, we conduct the control updates simultaneously with the solution of adjoint equation after solving the state equation. 

Specifically, to discretize the update rule for control variables \eqref{Parallel-ODE-Weights-Update-AL} for any $0\leq k\leq K-1$, \textit{i.e.},
\begin{equation*}
	w_t^k \leftarrow w_t^k - \eta \left( p_t^{k} \frac{\partial f(x_t^k,w^k_t)}{\partial w}\right) \ \ \ \textnormal{on} \ [s_k,s_{k+1}],
\end{equation*}
we adopt the numerical scheme \eqref{ResNet-Parameter-Updates} to guarantee the accurate gradient information \cite{gholami2019anode}, that is,
\begin{equation}
\begin{split}
	W_{kn+m}^k & \leftarrow W_{kn+m}^k - \eta \left( P^k_{kn+m+1} \frac{\partial F(X^k_{kn+m},W^k_{kn+m})}{\partial W} \right) \\
	& = W_{kn+m}^k - \eta \left(  (1-\delta) \left( \beta \frac{\partial \psi(\lambda_{k+1},X^k_{kn+n})}{\partial X^k_{kn+m+1}} + \kappa_{k+1} \frac{\partial X^k_{kn+n}}{\partial X^k_{kn+m+1}} \right) + \delta \frac{\partial \varphi(X_{kn+n}^{k})}{\partial X^{k}_{kn+m+1}} \right) \frac{\partial X^k_{kn+m+1}}{\partial W^k_{kn+m}} \\
	& = W^k_{kn+m} - \eta \bigg(  (1-\delta) \left( \beta \frac{\partial \psi(\lambda_{k+1},X^k_{kn+n})}{\partial W^k_{kn+m}} + \kappa_{k+1} \frac{\partial X^k_{kn+n}}{\partial W^k_{kn+m}} \right) + \delta \frac{\partial \varphi(X_{kn+n}^{k})}{\partial W^k_{kn+m}} \bigg)
\end{split}	
\label{Parallel-ResNet-Backpropagation}
\end{equation}
where the second equality holds by \eqref{Parallel-ResNet-Feature-Flow} and \eqref{Parallel-ResNet-Gradient-Propagation}. Next, we have by \eqref{Parallel-ODE-Slack-Variables-Update-AL} and \eqref{Parallel-ResNet-Gradient-Propagation} that the correction of auxiliary variables satisfies $\lambda_0 \equiv x_0$ and
\begin{equation}
	 \lambda_k \leftarrow \lambda_k - \eta \left( \beta \frac{\partial \psi(\lambda_k,X^{k-1}_{kn})}{\partial \lambda} + (1-\delta) \left( \beta \frac{\partial \psi(\lambda_{k+1},X^k_{kn+n})}{\partial X^k_{kn}} + \kappa_{k+1}\frac{\partial X^k_{kn+n}}{\partial X^k_{kn}} \right) + \delta \frac{\partial \varphi(X_{kn+n}^{k})}{\partial X^{k}_{kn}} - \kappa_k \right)  
	\label{Parallel-ResNet-Slack-Variables-Update}
\end{equation}
for $1\leq k\leq K-1$, while the update rule \eqref{Parallel-ODE-Multiplier-Update-AL} for Lagrangian multiplier $\kappa_k$ is given by
\begin{equation}
	\kappa_0 = 0\ \qquad \textnormal{and}\ \qquad \kappa_k \leftarrow \kappa_k - \frac{\eta}{2\beta}\left( \lambda_k - X^{k-1}_{kn} \right)
	\label{Parallel-ResNet-Multiplier-Update}
\end{equation}
for $1\leq k\leq K-1$. Clearly, operations \eqref{Parallel-ResNet-Slack-Variables-Update} and \eqref{Parallel-ResNet-Multiplier-Update} require communication between adjacent layers which can impede the performance of parallel computations.

Consequently, the layer-parallel training approach for solving \eqref{ResNet-Training-Task} can be formulated as the sequential operations
\begin{tcolorbox}[
						  standard jigsaw,
    						  opacityback=0,  
						]
\begin{empheq}{align*}
	\bullet \ \textnormal{decoupled forward pass and backpropagation}\ \eqref{Parallel-ResNet-Feature-Flow}, \eqref{Parallel-ResNet-Backpropagation} \qquad \bullet \ \textnormal{global communication}\ \eqref{Parallel-ResNet-Slack-Variables-Update},\eqref{Parallel-ResNet-Multiplier-Update}
\end{empheq}
\end{tcolorbox}
at each iteration, which breaks the forward, backward and update locking issues \cite{jaderberg2017decoupled}. 

\section{Initialization of Algorithm 1} \label{appendix-initialization}

\begin{algorithm2e}[htp]
\SetAlgoLined
\SetNlSty{texttt}{[\!}{\!]}
\SetAlgoNlRelativeSize{0}
\SetNlSkip{0em}
\tcp{Initialization.}
\nl divide the ResNet model into $K$ local sub-models (\textit{e.g.}, the uniform decomposition depicted in \autoref{fig-multifidelity-forward-propagation})\; 
\nl generate the initial guess of parameters and auxiliary variables (\textit{e.g.}, copy from a ResNet trained with one epoch)\;
\nl use a proper metric (\textit{e.g.}, squared $\ell_2$-norm) and coefficients (\textit{e.g.}, an increasing sequence) for penalty function\;
\nl set Lagrangian multipliers to zero; \tcp{degenerate to penalty method if $\kappa_k\equiv 0$ hereafter}
\nl schedule proper learning rates for network parameters $\eta_w$, auxiliary variables $\eta_\lambda$, and multipliers $\eta_\kappa$\;

\caption{Initialization of the Layer-parallel Training Algorithm}
\end{algorithm2e}


\section{Additional Experiments}\label{appendix-experiments}

\begin{figure*}[htb]

\centering
\minipage{0.32\textwidth}
  \includegraphics[width=\linewidth]{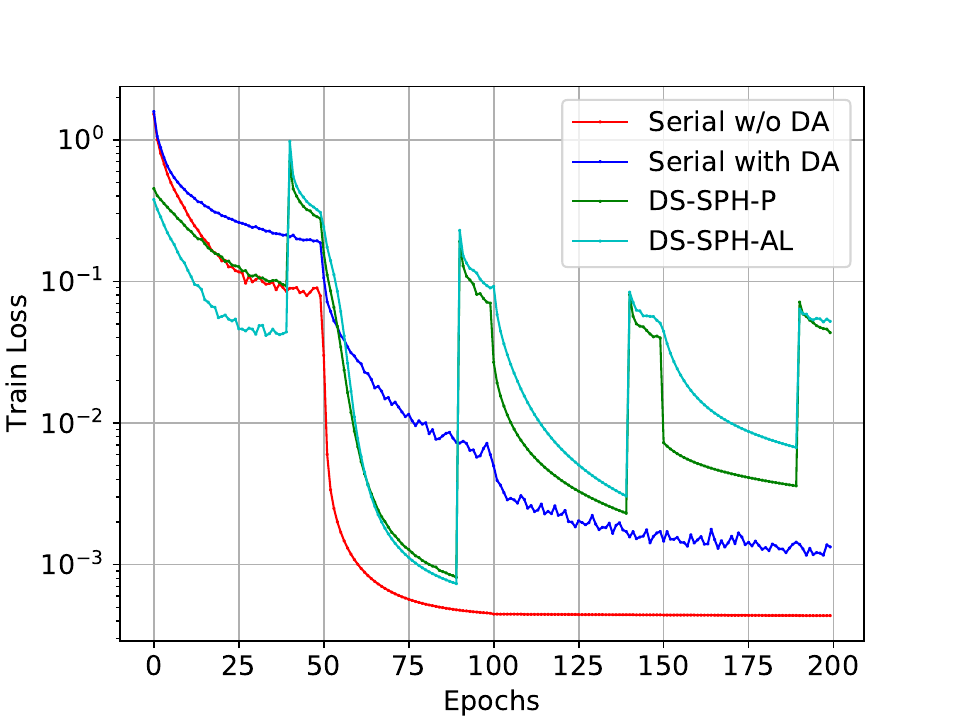}
\endminipage
\hfill
\minipage{0.32\textwidth}
  \includegraphics[width=\linewidth]{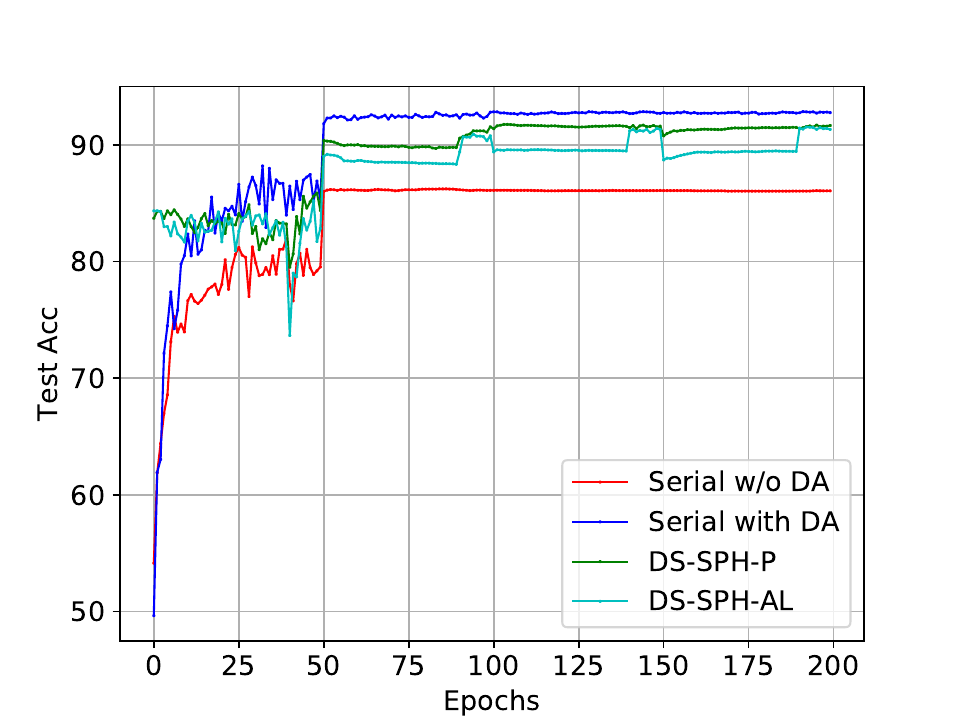}
\endminipage
\hfill
\minipage{0.32\textwidth}
  \includegraphics[width=\linewidth]{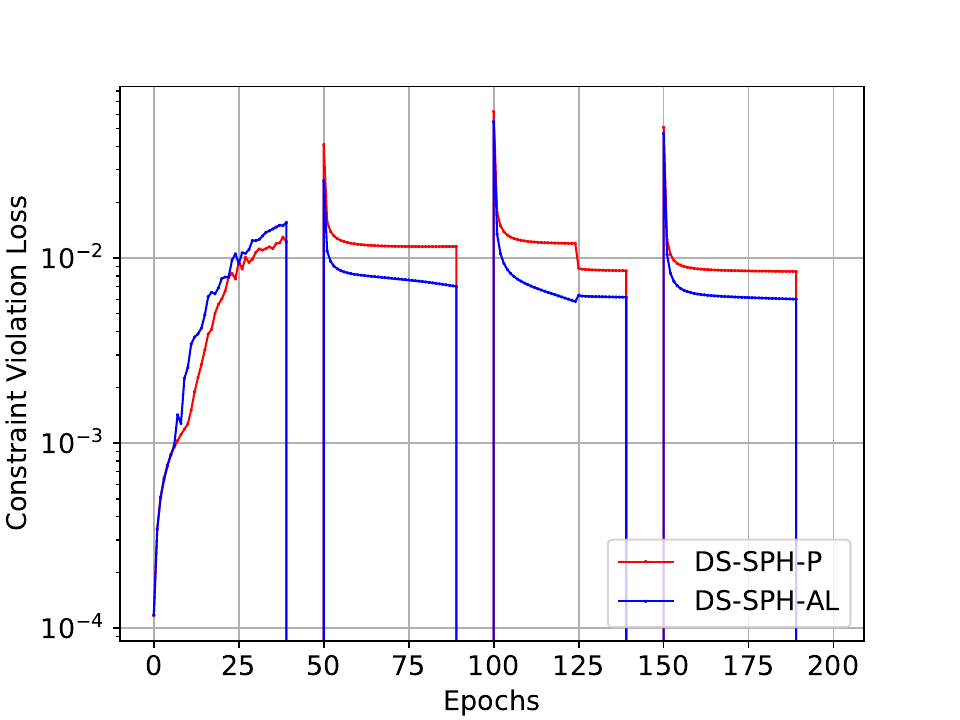}
\endminipage

\vspace{-0.08cm}

\caption{Training loss through the full serial forward pass, testing loss and constraint violation for ResNet-110 on CIFAR-10 dataset, where $K=3$ and $\gamma_H = 1/4$ (the serial and parallel portions are executed alternatively).}
\label{Fig-Learning-Curves-cifar10-K-3-SPP}

\vspace{0.1cm}
\end{figure*}

\begin{figure*}[htb]

\centering
\minipage{0.32\textwidth}
  \includegraphics[width=\linewidth]{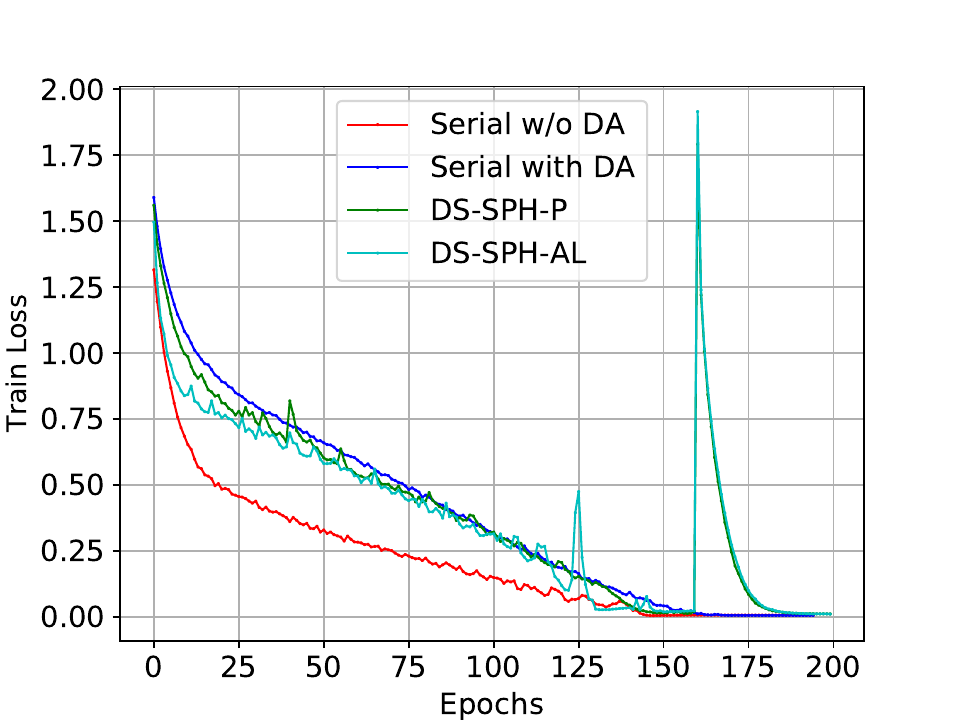}
\endminipage
\hfill
\minipage{0.32\textwidth}
  \includegraphics[width=\linewidth]{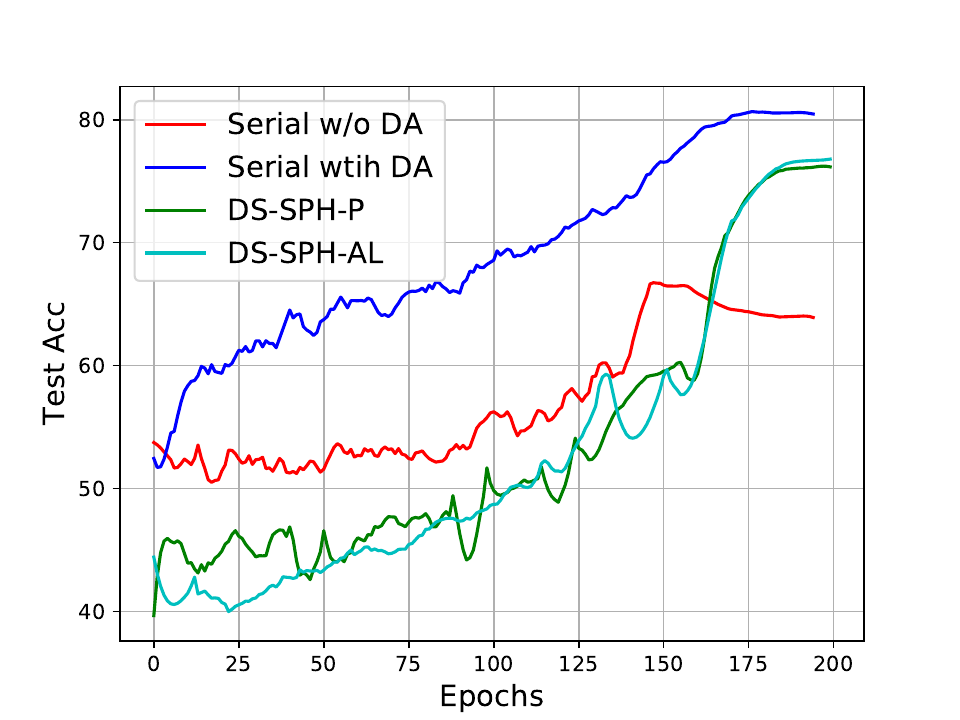}
\endminipage
\hfill
\minipage{0.32\textwidth}
  \includegraphics[width=\linewidth]{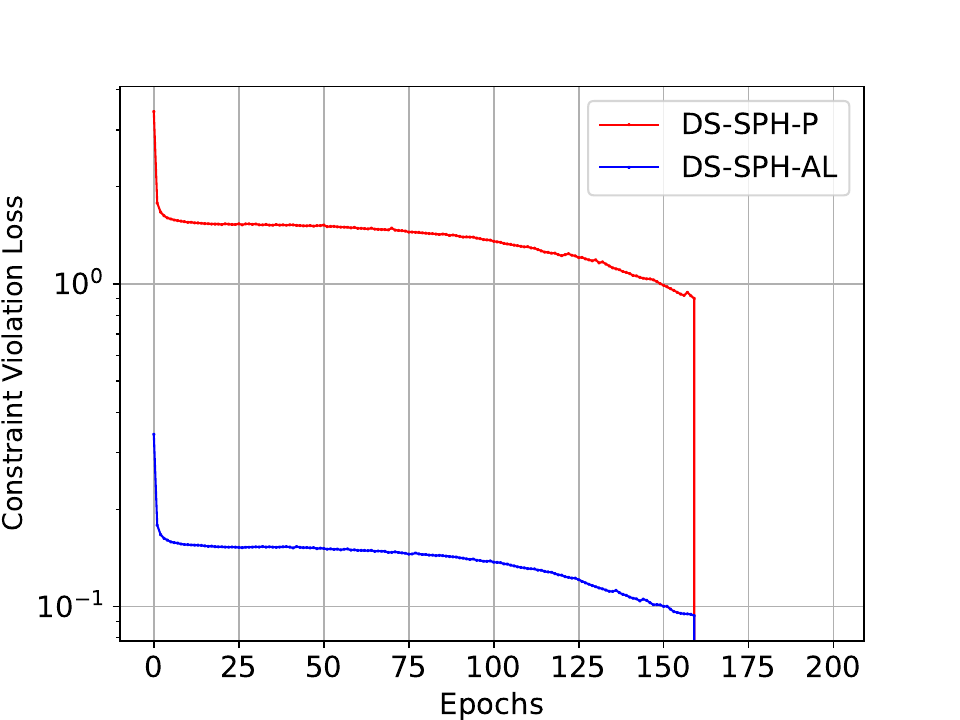}
\endminipage

\vspace{-0.08cm}

\caption{Training loss through the full serial forward pass, testing loss and constraint violation for WideResNet on CIFAR-100 dataset, where $K=3$ and $\gamma_H = 1/4$ (the parallel portion is first executed, followed by the serial portion to finish the training).}
\label{Fig-Learning-Curves-cifar100-K-3-SPP}

\vspace{0.1cm}
\end{figure*}

 \textbf{Learning curves.} We show in \autoref{Fig-Learning-Curves-cifar10-K-3-SPP} and \autoref{Fig-Learning-Curves-cifar100-K-3-SPP} the learning curves of ResNet on CIFAR-10 and WideResNet on CIFAR-100, respectively. As can be seen from \autoref{Fig-Learning-Curves-cifar10-K-3-SPP} (right) and \autoref{Fig-Learning-Curves-cifar100-K-3-SPP} (right), a large jump of constraint violation appears when the training is switched from parallel to serial, which helps improve the network parameters together with the use of data augmentation (see \autoref{Fig-Learning-Curves-cifar10-K-3-SPP} (left and middle) and \autoref{Fig-Learning-Curves-cifar100-K-3-SPP} (left and middle)). Moreover, by using the same penalty coefficient, the constraint violation associated with AL method is smaller than that of penalty method (see \autoref{Fig-Learning-Curves-cifar10-K-3-SPP} (right) and \autoref{Fig-Learning-Curves-cifar100-K-3-SPP}(right)), which validates our theoretical analysis \eqref{Constraint-Violation-Penalty} and \eqref{Constraint-Violation-AL}.


\end{document}